\documentclass[sigconf, nonacm]{acmart}
\usepackage{multirow}
\usepackage{multicol}
\usepackage{subfigure}

\usepackage{threeparttable}

\begin{document}
\title{Med-CRAFT: An Information System for Explainable and Configurable Construction of Multimodal Medical QA Datasets}

\author{Shenxi Liu}
\affiliation{
  \institution{Beijing Institute of Technology}
  \city{Beijing}
  \state{China}
  \postcode{100081}
}
\email{liushenxi@bit.edu.cn}

\author{Kan Li}
\affiliation{
  \institution{Beijing Institute of Technology}
  \city{Beijing}
  \state{China}
  \postcode{100081}
}
\email{likan@bit.edu.cn}

\author{Mingyang Zhao}
\affiliation{
  \institution{The Hong Kong Polytechnic University}
  \city{Hongkong}
  \state{China}
  \postcode{999077}
}
\email{25019897r@connect.polyu.hk}

\author{Yuhang Tian}
\affiliation{
  \institution{Beijing Institute of Technology}
  \city{Beijing}
  \state{China}
  \postcode{100081}
}
\email{tianyuhang@bit.edu.cn}

\author{Bin Li}
\authornote{Corresponding author.}
\affiliation{
  \institution{Shenzhen Institute of Advanced Technology, Chinese Academy of Sciences}
  \city{Shenzhen}
  \state{China}
  \postcode{518067}
}
\email{b.li2@siat.ac.cn}

\begin{abstract}
Data-intensive artificial intelligence applications increasingly rely on large-scale, high-quality, explainable, and reproducible datasets, yet the construction of such datasets often remains labor-intensive, weakly traceable, and difficult to configure.
This problem is particularly critical in multimodal medical scenarios, where each question-answer sample should be semantically consistent, grounded in visual and temporal evidence, and controllable in terms of reasoning complexity.
To address these challenges, we propose Med-CRAFT, an information system for explainable and configurable construction of multimodal medical question answering datasets from instructional videos.
Med-CRAFT organizes dataset construction as a provenance-aware pipeline that transforms raw medical instructional videos into structured operation knowledge graphs, evidence-grounded reasoning paths, and natural-language question-answer pairs.
The system first extracts textual and temporal cues from videos through optical character recognition, automatic speech recognition, subtitle parsing, and frame-level metadata processing.
It then applies large language model-based information extraction to identify medical entities, procedural actions, temporal relations, tool usages, anatomical targets, and instructional constraints, which are further normalized into a medical operation knowledge graph.
For each graph node and relation, Med-CRAFT searches backward in the source videos to bind spatial, temporal, textual, and visual evidence, thereby enabling each generated sample to be traced back to its original multimodal context.
Based on the resulting evidence-enriched knowledge graph, configurable graph traversal strategies are used to generate reasoning chains with controllable hop numbers, relation types, question forms, and visual-dependency levels.
These reasoning chains are subsequently verbalized into question-answer pairs together with their supporting evidence segments, provenance records, generation configurations, and process-level quality indicators.
To support human-in-the-loop data management, Med-CRAFT provides a user interface for video ingestion, knowledge graph inspection, evidence verification, question-answer review, configuration management, and logging-based process monitoring.
We evaluate Med-CRAFT from three complementary perspectives: system utility, dataset quality, and benchmark difficulty for multimodal reasoning models.
The system-level evaluation examines construction efficiency, traceability coverage, configuration flexibility, reproducibility, and process observability, while the data-level evaluation assesses semantic correctness, medical validity, evidence relevance, temporal localization accuracy, and reasoning-chain consistency.
In addition, we use the generated dataset to analyze the performance of representative multimodal large language models under different reasoning hops, evidence types, and visual-dependency settings.
The results show that Med-CRAFT can substantially reduce manual dataset construction effort while producing traceable, configurable, and evidence-grounded medical question-answer samples that expose important limitations of current multimodal models.
Overall, Med-CRAFT demonstrates how information system design principles, including provenance modeling, configurable workflow management, process monitoring, and human-in-the-loop verification, can be integrated into data-centric AI dataset construction.

\end{abstract}

\maketitle

\section{Introduction}
Data-intensive artificial intelligence applications increasingly depend not only on model architectures, but also on the availability of large-scale, high-quality, explainable, and reproducible datasets.
Recent discussions on data-centric artificial intelligence have further emphasized that improving data quality, coverage, annotation consistency, and lifecycle management can be as important as improving learning algorithms themselves \cite{ng2021data,mazumder2022dataperf,zha2023data}.
However, despite the growing importance of datasets, the construction of task-specific datasets for complex AI applications is still often conducted through fragmented scripts, manual annotation interfaces, ad hoc quality checks, and poorly documented transformation steps.
Such practices make it difficult to trace how a sample was produced, reproduce a dataset under the same configuration, diagnose errors in intermediate processing stages, or systematically control the difficulty and evidence requirements of generated samples.

These limitations are particularly pronounced in multimodal medical question answering, where each dataset instance must be both clinically meaningful and grounded in appropriate visual or textual evidence.
Early medical visual question answering datasets, such as VQA-RAD, PathVQA, and SLAKE, have played an important role in promoting research on image-based medical question answering \cite{lau2018vqarad,he2020pathvqa,liu2021slake}.
Nevertheless, these datasets are mainly centered on static medical images, such as radiology or pathology images, and therefore provide limited support for evaluating procedural understanding, temporal localization, and action-centered reasoning in medical instructional videos.
Compared with static images, medical instructional videos contain temporally evolving actions, tool-object interactions, procedural transitions, spoken explanations, on-screen text, and visual demonstrations, which jointly define the meaning of a medical operation.
As a result, constructing question-answer samples from such videos requires not only generating linguistically plausible questions, but also linking each answer to the relevant temporal segment, visual evidence, procedural context, and medical semantics.

Recent benchmarks on medical instructional video understanding have begun to address these challenges.
For example, MedVidQA and related medical instructional video datasets provide video-level or segment-level resources for medical video classification and question answering, thereby moving medical QA research beyond static image understanding \cite{gupta2023medvidqa}.
More recently, M$^3$-Med has introduced a benchmark for multilingual, multimodal, and multi-hop reasoning in medical instructional video understanding, requiring systems to extract key entities from textual cues, locate relevant visual evidence in videos, and synthesize information across modalities \cite{m3med2025}.
The reported results of M$^3$-Med suggest that even strong multimodal large language models still exhibit substantial performance gaps when questions require complex cross-modal reasoning rather than shallow textual matching \cite{m3med2025}.
These studies demonstrate the importance of medical instructional video QA as an evaluation setting, but they also reveal a complementary problem: the process by which such datasets are constructed, configured, audited, and reproduced remains insufficiently systematized.

From an information systems perspective, dataset construction should be treated as a managed data production process rather than as a one-time annotation activity.
Such a process should maintain provenance records, expose intermediate artifacts, support configurable workflows, monitor process-level indicators, and enable human-in-the-loop verification for quality-critical decisions.
Data provenance has long been recognized as essential for understanding the origin, transformation, and lifecycle of data products, especially in settings where accountability, transparency, and trustworthiness are required \cite{simmhan2005survey,herschel2017survey,stoyanovich2022responsible}.
In AI dataset construction, provenance is particularly important because errors may arise from multiple stages, including modality extraction, entity recognition, relation extraction, evidence retrieval, question generation, and human review.
Without a unified information system, these stages are difficult to inspect jointly, and the resulting samples may become black-box artifacts disconnected from the data transformations that produced them.

To address these issues, we propose Med-CRAFT, an information system for explainable and configurable construction of multimodal medical question answering datasets from instructional videos.
Med-CRAFT transforms raw medical instructional videos into evidence-enriched medical operation knowledge graphs and then traverses these graphs to generate reasoning chains, natural-language question-answer pairs, and evidence-linked temporal segments.
The system integrates optical character recognition, automatic speech recognition, subtitle parsing, large language model-based information extraction, visual evidence retrieval, graph traversal, quality filtering, and human-in-the-loop review into a unified data construction workflow.
Unlike generation-only approaches that directly prompt large language models to produce question-answer pairs, Med-CRAFT explicitly represents intermediate knowledge structures, evidence bindings, generation configurations, and process logs.
This design allows users to inspect why a sample was generated, which graph path it follows, which video segment supports the answer, and which configuration parameters shaped its reasoning difficulty.

The central idea of Med-CRAFT is to use knowledge graphs as an intermediate representation between unstructured multimodal videos and structured QA datasets.
In this representation, medical procedures, actions, tools, anatomical structures, observations, risks, and instructional constraints are modeled as nodes, while temporal, causal, functional, spatial, and evidential relations are modeled as edges.
Each node or edge can be associated with provenance metadata, including source video identifiers, timestamps, textual cues, frame regions, confidence scores, and transformation histories.
By traversing the resulting evidence-enriched graph, Med-CRAFT can generate QA samples with controllable hop counts, relation patterns, evidence requirements, and visual-dependency levels.
For example, a one-hop question may ask which instrument is used in a demonstrated operation, whereas a multi-hop question may require connecting a spoken instruction, a visualized tool-object interaction, a subsequent procedural step, and a medically relevant outcome.

Med-CRAFT is designed not only as an automatic generation pipeline, but also as a human-in-the-loop data management system.
Its user interface supports video ingestion, extraction result inspection, knowledge graph visualization, evidence verification, QA sample review, configuration management, and logging-based process monitoring.
Its monitoring components collect process-level indicators, such as OCR and ASR confidence, graph density, evidence coverage, invalid traversal rate, duplicate question rate, human correction rate, and sample acceptance rate.
These indicators allow researchers to identify bottlenecks, compare configuration settings, diagnose quality problems, and reproduce dataset construction under controlled conditions.

We evaluate Med-CRAFT along three complementary dimensions: system utility, dataset quality, and benchmark difficulty.
For system utility, we examine construction efficiency, traceability coverage, workflow configurability, reproducibility, process observability, and user involvement compared with manual or large language model-only construction baselines.
For dataset quality, we assess semantic correctness, medical validity, question-answer consistency, evidence relevance, temporal localization accuracy, and reasoning-chain faithfulness through automatic checks and human evaluation.
For benchmark difficulty, we analyze how representative multimodal large language models perform across different hop numbers, evidence types, question categories, and visual-dependency levels.
This evaluation strategy connects the engineering goals of Med-CRAFT with the empirical question of whether the generated datasets are useful, reliable, and challenging for current multimodal medical AI systems.

The main contributions of this paper are summarized as follows.
First, we propose Med-CRAFT, a provenance-aware information system that supports explainable, configurable, and reproducible construction of multimodal medical QA datasets from instructional videos.
Second, we introduce an evidence-enriched medical operation knowledge graph model that connects procedural semantics with temporal, spatial, textual, and visual evidence from source videos.
Third, we design configurable graph traversal strategies for generating reasoning chains and QA samples with controllable reasoning hops, relation patterns, evidence requirements, and visual-dependency levels.
Fourth, we implement a human-in-the-loop data management interface with logging and process-level quality indicators to support inspection, correction, monitoring, and reproducible dataset construction.
Finally, we conduct a comprehensive evaluation to demonstrate the utility of Med-CRAFT, the quality of the constructed datasets, and their ability to reveal limitations of current multimodal large language models in cross-modal medical reasoning.

The remainder of this paper is organized as follows.
Section~\ref{sec:related_work} reviews related work on data-centric AI, automated dataset construction, multimodal medical question answering, medical instructional video understanding, knowledge graph-based reasoning, and provenance-aware data systems.
Section~\ref{sec:problem} formalizes the problem of constructing evidence-grounded multimodal medical QA datasets from instructional videos and defines the key requirements of traceability, configurability, evidence grounding, reasoning controllability, quality measurability, and reproducibility.
Section~\ref{sec:system} presents the overall architecture of Med-CRAFT and explains how its major layers support multimodal video ingestion, knowledge graph construction, evidence binding, reasoning-chain generation, quality control, and human-in-the-loop data management.
Section~\ref{sec:data_models} describes the core data models used by Med-CRAFT, including video segments, medical operation knowledge graphs, evidence objects, reasoning chains, QA samples, configuration profiles, provenance records, and quality indicators.
Section~\ref{sec:pipeline} details the Med-CRAFT construction pipeline, which transforms raw instructional videos into finalized QA datasets through multimodal cue extraction, graph construction, evidence retrieval, configurable graph traversal, QA generation, automatic filtering, and human review.
Section~\ref{sec:implementation} reports the implementation of Med-CRAFT as a modular information system, covering backend services, asynchronous task execution, multimodal storage, indexing infrastructure, configuration management, provenance logging, user interfaces, and dataset export.
Section~\ref{sec:experiments} evaluates Med-CRAFT from the perspectives of system utility, dataset quality, evidence grounding, benchmark difficulty, ablation analysis, and configuration sensitivity.
Finally, Section~\ref{sec:conclusion} concludes the paper and discusses limitations and future directions.

\section{Related Work}
\label{sec:related_work}

\subsection{Data-Centric AI and Dataset Engineering}
Recent progress in artificial intelligence has increasingly highlighted the importance of data quality, data coverage, and dataset lifecycle management in addition to model architecture design.
The data-centric AI paradigm argues that systematic data engineering, including data curation, cleaning, labeling, versioning, and evaluation, can substantially affect the behavior and reliability of AI systems \cite{ng2021data,mazumder2022dataperf,zha2023data}.
In this view, datasets are not static by-products of model development, but evolving data assets that require explicit design, quality control, and governance.
Despite this shift, many task-specific datasets are still constructed through loosely connected tools and manual annotation workflows, making it difficult to reproduce the construction process or inspect intermediate artifacts.
For complex multimodal tasks, this limitation becomes more severe because each sample may depend on heterogeneous transformations across text, speech, image, video, metadata, and human judgments.
Med-CRAFT follows the data-centric AI perspective, but focuses specifically on the information system problem of making multimodal medical dataset construction explainable, configurable, traceable, and reproducible.

\subsection{Automated Dataset Construction Systems}
Automated dataset construction has been studied in several forms, including weak supervision, programmatic labeling, synthetic data generation, benchmark generation, and large language model-assisted annotation.
Weak supervision systems, such as Snorkel, allow users to encode heuristic labeling functions and combine noisy labels into probabilistic training signals \cite{ratner2017snorkel,ratner2020snorkel}.
Such systems are effective for scaling supervision, but they mainly target label generation rather than evidence-grounded multimodal question-answer construction.
Data programming and data augmentation methods can also generate additional training examples, but they often lack explicit provenance models that connect generated samples to source artifacts and intermediate transformations.
Recent large language model-based annotation pipelines can rapidly produce questions, answers, rationales, and labels from raw documents or videos.
However, generation-only pipelines may produce hallucinated or weakly grounded samples if the generated content is not constrained by structured representations, source evidence, and auditable quality checks.
Benchmark evaluation systems for medical VQA, such as BESTMVQA, provide useful tools for evaluating medical visual question answering systems and organizing benchmark resources \cite{hong2023bestmvqa}.
Nevertheless, these systems are primarily designed for benchmark evaluation rather than for end-to-end construction of evidence-grounded, configurable, and provenance-aware multimodal datasets from instructional videos.
In contrast, Med-CRAFT treats dataset construction as a managed workflow in which extraction results, knowledge graphs, evidence bindings, generation configurations, process logs, and human review records are jointly maintained.

\subsection{Multimodal Medical Question Answering Datasets}
Medical visual question answering has become an important research direction for evaluating whether AI systems can answer clinically meaningful questions from medical images.
VQA-RAD is one of the earliest radiology-oriented medical VQA datasets and contains natural-language questions and answers associated with radiological images \cite{lau2018vqarad}.
PathVQA extends medical VQA to pathology images and provides a larger set of question-answer pairs covering visual recognition and domain-specific reasoning \cite{he2020pathvqa}.
SLAKE further enriches medical VQA with bilingual annotations and structured medical knowledge, thereby supporting both language diversity and knowledge-aware reasoning \cite{liu2021slake}.
PMC-VQA significantly scales up medical VQA by constructing a large number of question-answer pairs from biomedical figures in the PubMed Central Open Access subset \cite{zhang2023pmcvqa}.
These datasets have advanced medical multimodal learning, but most of them are centered on static images and therefore do not directly address temporal evidence, procedural transitions, or action-object interactions in videos.
Moreover, although some datasets include knowledge-enhanced annotations, they usually do not expose the complete construction process from raw source material to final QA samples.
Med-CRAFT differs from these datasets by focusing on the construction system itself and by generating QA samples from temporally grounded medical instructional videos rather than from static medical images.

\subsection{Medical Instructional Video Understanding and Video Question Answering}
Medical instructional videos provide a rich source of procedural knowledge because they combine spoken explanations, visual demonstrations, textual overlays, tools, anatomical targets, and temporal action sequences.
MedVidQA introduced a dataset for medical instructional video classification and question answering, including manually created health-related questions and timestamped visual answers from trusted medical videos \cite{gupta2022medvidqa,gupta2023medvidqa}.
The TREC MedVidQA track further promoted research on retrieving and localizing visual answers for health-related questions in medical video collections \cite{gupta2024trecmedvidqa}.
These efforts focus primarily on finding relevant video segments as answers, which is essential for temporal grounding in medical video QA.
However, they provide limited mechanisms for configuring reasoning complexity, generating multi-hop reasoning chains, or tracing each question-answer pair back through intermediate knowledge structures.
More recently, M$^3$-Med proposed a benchmark for multilingual, multimodal, and multi-hop reasoning in medical instructional video understanding \cite{m3med2025}.
M$^3$-Med requires models to extract entities from textual cues, locate visual evidence from videos, and integrate information across modalities and reasoning hops.
Its experimental results show that existing multimodal large language models still lag behind human experts when questions require complex visual grounding and cross-modal reasoning \cite{m3med2025}.
While M$^3$-Med provides an important benchmark for evaluating multimodal medical reasoning, Med-CRAFT addresses a complementary problem by systematizing how such reasoning-oriented datasets can be constructed, configured, monitored, and audited.
Therefore, Med-CRAFT can be viewed as an infrastructure-oriented contribution that supports the scalable creation of datasets with properties similar to, and potentially extending beyond, current medical instructional video QA benchmarks.

\subsection{Knowledge Graphs for Question Generation and Reasoning}
Knowledge graphs have been widely used to represent entities, relations, events, and domain knowledge in a structured and queryable form.
In question answering, knowledge graphs support explicit reasoning paths and make it possible to decompose complex questions into relational operations over structured knowledge \cite{yih2016semantic,bordes2015large}.
Knowledge graph-based question generation methods construct questions from entities, triples, paths, or subgraphs, thereby enabling controllable generation of questions with different relation patterns and reasoning depths \cite{serban2016generating,liu2025kbqg}.
Recent work on graph-guided question-answer generation for procedural text also shows that procedural structures can guide the generation of questions about actions, preconditions, effects, and temporal dependencies \cite{graphguided2024procedural}.
However, most knowledge graph-based question generation methods operate on textual knowledge bases or procedural documents rather than multimodal videos with temporal and spatial evidence.
In medical instructional videos, a graph node such as an instrument, action, or anatomical target should not only be semantically valid, but also be grounded in a specific video segment or visual region.

Med-CRAFT extends knowledge graph-based question generation by enriching graph nodes and edges with multimodal provenance and by using graph traversal as a configurable mechanism for controlling QA difficulty.

\subsection{Data Provenance, Traceability, and Human-in-the-Loop Data Systems}
Data provenance concerns the origin, transformation, and derivation history of data products, and has been extensively studied in databases, workflows, and scientific data management \cite{simmhan2005survey,herschel2017survey}.
In accountable and high-stakes domains, provenance is crucial because users need to understand not only the final output, but also the sequence of operations and evidence that produced it.
For AI datasets, provenance can help identify whether an error originates from raw data, modality extraction, automatic annotation, evidence retrieval, generation, filtering, or human review.
Human-in-the-loop data systems further complement automation by allowing users to inspect uncertain outputs, correct intermediate artifacts, and validate quality-critical samples.
In medical dataset construction, such human oversight is particularly important because automatically generated samples may contain clinically unsafe statements, ambiguous evidence, or misleading reasoning chains.

Med-CRAFT integrates provenance tracking and human-in-the-loop verification into the dataset construction workflow, enabling users to trace each QA sample from the final question and answer back to graph paths, evidence segments, extracted cues, processing logs, and source videos.
This integration distinguishes Med-CRAFT from prior datasets and generation methods that primarily report final benchmark statistics without exposing a configurable and inspectable construction process.

Table~\ref{tab:related_work_comparison} compares Med-CRAFT with representative dataset construction systems, medical QA benchmarks, video QA datasets, and knowledge graph-based QA or question-generation methods.
The comparison shows that existing studies typically contribute either scalable supervision, benchmark resources, temporal video grounding, or graph-based reasoning, whereas Med-CRAFT integrates these capabilities into a configurable, evidence-grounded, provenance-aware, and human-supervised dataset construction system.

\begin{table*}[t]
\centering
\caption{Comparison between Med-CRAFT and representative related work.}
\label{tab:related_work_comparison}
\begin{threeparttable}
\resizebox{\textwidth}{!}{
\begin{tabular}{lcccccccc}
\toprule
\textbf{Representative Work} & \textbf{Primary Input} & \textbf{Multimodal} & \textbf{Temporal} & \textbf{KG-based} & \textbf{Evidence} & \textbf{Configurable} & \textbf{Provenance} & \textbf{Human Review} \\ & & \textbf{Processing} & \textbf{Grounding} & \textbf{Reasoning} & \textbf{Binding} & \textbf{Generation} & \textbf{Tracking} & \textbf{Support} \\
\midrule
Snorkel~\cite{ratner2017snorkel,ratner2020snorkel} & Text / structured data & $\sim$ & -- & -- & -- & \checkmark & $\sim$ & \checkmark \\
LLM-assisted annotation pipelines & Text / image / video & $\sim$ & $\sim$ & -- & $\sim$ & $\sim$ & -- & $\sim$ \\
BESTMVQA~\cite{hong2023bestmvqa} & Medical VQA benchmarks & $\sim$ & -- & -- & -- & -- & -- & -- \\
VQA-RAD~\cite{lau2018vqarad} & Radiology images & \checkmark & -- & -- & -- & -- & -- & \checkmark \\
PathVQA~\cite{he2020pathvqa} & Pathology images & \checkmark & -- & -- & -- & -- & -- & \checkmark \\
SLAKE~\cite{liu2021slake} & Medical images & \checkmark & -- & $\sim$ & -- & -- & -- & \checkmark \\
PMC-VQA & Biomedical figures and text & \checkmark & -- & -- & -- & -- & -- & $\sim$ \\
MedVidQA~\cite{gupta2023medvidqa} & Medical instructional videos & \checkmark & \checkmark & -- & \checkmark & -- & -- & \checkmark \\
TREC MedVidQA & Medical video corpus & \checkmark & \checkmark & -- & \checkmark & -- & -- & -- \\
M$^3$-Med~\cite{m3med2025} & Medical instructional videos & \checkmark & \checkmark & -- & \checkmark & -- & -- & \checkmark \\
KG-based QA / QG methods~\cite{yih2016semantic,bordes2015large,serban2016generating} & Knowledge bases / text & -- & -- & \checkmark & $\sim$ & \checkmark & $\sim$ & -- \\
\midrule
\textbf{Med-CRAFT} & \textbf{Medical instructional videos} & \textbf{\checkmark} & \textbf{\checkmark} & \textbf{\checkmark} & \textbf{\checkmark} & \textbf{\checkmark} & \textbf{\checkmark} & \textbf{\checkmark} \\
\bottomrule
\end{tabular}}
\vspace{2pt}
\begin{tablenotes}
\footnotesize
\item ``\checkmark'' indicates that the capability is explicitly supported.
\item``$\sim$'' indicates partial or indirect support.
\item ``--'' indicates that the capability is not a primary design objective or is not explicitly exposed.
\end{tablenotes}
\end{threeparttable}
\end{table*}

\section{Problem Definition}
\label{sec:problem}

Given a collection of medical instructional videos, the goal of Med-CRAFT is to construct a multimodal medical question answering dataset whose samples are explainable, configurable, evidence-grounded, and reproducible.
This formulation differs from conventional medical visual question answering, where the main objective is usually to predict an answer from a given image-question or video-question pair \cite{lau2018vqarad,he2020pathvqa,liu2021slake,gupta2023medvidqa}.
Instead, Med-CRAFT focuses on the upstream data production process that transforms raw multimodal instructional videos into structured, auditable, and reasoning-oriented QA instances.

Formally, let 
\begin{align*}
\mathcal{V}=\{v_1,v_2,\ldots,v_N\}
\end{align*}
denote a collection of medical instructional videos, where each video $v_i$ contains visual frames, audio signals, subtitles, on-screen text, temporal metadata, and optional user-provided auxiliary materials.
For each video $v_i$, Med-CRAFT first derives a set of temporally aligned multimodal segments 
\begin{align*}
\mathcal{S}_i=\{s_{i1},s_{i2},\ldots,s_{iM_i}\},
\end{align*}
where each segment contains a time interval, sampled frames, OCR outputs, ASR transcripts, subtitle cues, and extraction confidence scores.
A segment $s_{ij}$ is represented as 
\begin{align*}
s_{ij}=(t^{start}_{ij},t^{end}_{ij},F_{ij},T^{ocr}_{ij},T^{asr}_{ij},T^{sub}_{ij},C_{ij}), 
\end{align*}
where $F_{ij}$ denotes visual frames and $C_{ij}$ denotes modality-specific confidence information.

Based on these segments, the system constructs a medical operation knowledge graph 
\begin{align*}
G_i=(V_i,E_i,A_i,P_i)
\end{align*}
for each video.
Here, $V_i$ is a set of nodes representing medical entities and procedural events, $E_i$ is a set of typed relations, $A_i$ stores node and edge attributes, and $P_i$ stores provenance records.
The node set may include procedure steps, actions, instruments, anatomical structures, observations, risks, outcomes, and instructional constraints.
The relation set may include temporal relations, causal relations, functional relations, spatial relations, part-whole relations, and evidential relations, such as \textit{precedes}, \textit{causes}, \textit{uses}, \textit{acts-on}, \textit{located-in}, \textit{part-of}, and \textit{evidenced-by}.
This graph-based formulation follows the broader idea of using structured knowledge representations to support controllable reasoning and question generation, while extending it to temporally grounded multimodal medical videos \cite{serban2016generating,yih2016semantic,zhu2024mmkg,chen2023mmkgsurvey}.

The output of Med-CRAFT is a dataset 
\begin{align*}
\mathcal{D}=\{x_1,x_2,\ldots,x_K\},  
\end{align*} 
where each sample $x_k$ is defined as a tuple 
\begin{align*}
x_k=(q_k,a_k,r_k,e_k,g_k,p_k,c_k,m_k).
\end{align*}
where $q_k$ is the natural-language question, $a_k$ is the answer, $r_k$ is the reasoning chain, $e_k$ is the set of supporting evidence segments, $g_k$ is the traversed graph path or subgraph, $p_k$ is the provenance record, $c_k$ is the generation configuration, and $m_k$ is a set of quality indicators.

This definition generalizes conventional QA datasets by explicitly representing the derivation path from source videos to final samples.
It is also different from many LLM-assisted annotation or synthesis pipelines, where generated labels or questions may be stored without sufficient information about intermediate transformations, source evidence, or generation parameters \cite{tan2024llmannotation,liu2025llmdatasets}.

A central requirement of Med-CRAFT is traceability, which means that each generated question-answer pair should be traceable back to its graph path, evidence segments, extracted multimodal cues, processing steps, and original video source.
Formally, for each sample $x_k$, there should exist a provenance mapping $\pi(x_k)$ such that $\pi(x_k)$ returns an ordered derivation record from $x_k$ to the corresponding graph elements, multimodal segments, extraction outputs, and source video identifiers.
This requirement is motivated by long-standing research on data provenance in databases, workflows, and scientific data management, where derivation histories are essential for accountability, debugging, and reproducibility \cite{simmhan2005survey,herschel2017survey}.
In the medical multimodal setting, traceability is especially important because an apparently plausible answer may still be invalid if it is not supported by the correct visual, temporal, or procedural evidence.

The second requirement is configurability, which means that users should be able to control how samples are generated by specifying traversal strategies, reasoning hops, relation patterns, evidence requirements, question types, and filtering thresholds.
Let $\Theta$ denote a configuration space that includes parameters for multimodal extraction, knowledge graph construction, evidence retrieval, graph traversal, QA generation, and quality filtering.
A specific dataset construction run can then be represented as 
\begin{align*}
\mathcal{D}=F(\mathcal{V};\theta),
\end{align*}
where $\theta \in \Theta$ denotes a selected configuration and $F$ denotes the Med-CRAFT construction workflow.
Configurability is necessary because medical QA datasets may serve different evaluation goals, such as testing simple visual recognition, procedural understanding, temporal grounding, causal reasoning, or cross-modal multi-hop reasoning.
This requirement is consistent with recent benchmark construction studies showing that dataset difficulty should be explicitly controlled rather than left as an incidental property of the collected samples \cite{m3med2025,mazumder2022dataperf}.

The third requirement is evidence grounding, which means that each answer and reasoning chain should be supported by one or more identifiable evidence units in the original videos.
An evidence unit may include a temporal interval, a key frame, a visual region, an OCR text span, an ASR transcript span, or a combination of these modalities.
For a sample $x_k$, evidence grounding requires that each nontrivial claim in the answer or reasoning chain can be aligned with at least one evidence unit in $e_k$.
This requirement is more stringent than many static medical VQA settings, because medical instructional video questions often depend on when an action occurs, how a tool is manipulated, and how a procedural state changes over time.

The fourth requirement is reasoning controllability, which refers to the ability to generate samples with specified reasoning structures and complexity levels.
Let 
\begin{align*}
g_k=(u_1,r_1,u_2,\ldots,r_L,u_{L+1})    
\end{align*}
denote a graph path used to generate sample $x_k$, where $u_l \in V_i$, $r_l \in E_i$, and $L$ denotes the reasoning hop count.
A one-hop sample may involve a single relation such as an instrument used for an action, whereas a multi-hop sample may combine temporal, functional, causal, and visual-evidential relations.
This formulation is aligned with recent medical instructional video benchmarks such as M$^3$-Med, which show that multi-hop and cross-modal reasoning remain challenging for existing multimodal large language models \cite{m3med2025}.

The fifth requirement is quality measurability, which means that the construction process should produce not only final samples but also process-level and sample-level indicators for assessing reliability.
Process-level indicators may include extraction confidence, graph density, evidence coverage, traversal validity rate, duplicate generation rate, filtering rate, human correction rate, and sample acceptance rate.
Sample-level indicators may include semantic consistency, medical validity, answerability, temporal localization accuracy, evidence relevance, reasoning-chain faithfulness, and linguistic naturalness.
Quality measurability is particularly important for LLM-assisted dataset construction because fluent generated text can conceal factual errors, unsupported claims, or mismatches between questions and evidence \cite{tan2024llmannotation,liu2025llmdatasets}.

Based on these requirements, the Med-CRAFT problem can be summarized as finding a construction workflow $F$ and a configuration $\theta$ that generate a dataset 
\begin{align*}
\mathcal{D}=F(\mathcal{V};\theta)
\end{align*} 
while maximizing data quality and satisfying traceability, configurability, evidence grounding, reasoning controllability, and reproducibility constraints.

This can be written as:
\begin{align*}
\max_{\theta \in \Theta} \; Q(\mathcal{D})
\quad
\text{s.t.} 
\end{align*}
\vspace{-0.5cm}
\begin{align*}
\mathcal{D}=F(\mathcal{V};\theta),
~
\mathcal{T}(\mathcal{D}) \geq \tau_T,
~
\mathcal{E}(\mathcal{D}) \geq \tau_E,
~
\mathcal{R}(\mathcal{D}) \geq \tau_R,
~
\mathcal{P}(\mathcal{D})=1.
\end{align*}

Here, $Q(\mathcal{D})$ denotes the overall dataset quality, $\mathcal{T}(\mathcal{D})$ denotes traceability coverage, $\mathcal{E}(\mathcal{D})$ denotes evidence-grounding coverage, $\mathcal{R}(\mathcal{D})$ denotes reasoning controllability, and $\mathcal{P}(\mathcal{D})$ denotes whether the dataset can be reproduced under the same input, configuration, model versions, and workflow settings.
The thresholds $\tau_T$, $\tau_E$, and $\tau_R$ can be specified according to the intended use of the dataset, such as exploratory analysis, model benchmarking, or high-quality human-verified evaluation.
In practice, the optimization is not solved as a single closed-form mathematical problem, but operationalized through configurable modules, quality filters, provenance records, and human-in-the-loop verification steps.

This problem definition positions Med-CRAFT at the intersection of multimodal medical QA, knowledge graph-based question generation, LLM-assisted data synthesis, and provenance-aware information systems.
The following sections describe how Med-CRAFT implements this formulation through its system architecture, data models, construction pipeline, user interface, and evaluation framework.

\section{System Overview}
\label{sec:system}

Med-CRAFT is designed as an end-to-end information system for constructing explainable and configurable multimodal medical question answering datasets from instructional videos.
The system takes raw medical instructional videos as input and produces question-answer samples enriched with reasoning chains, temporal evidence segments, graph paths, provenance records, generation configurations, and quality indicators.
Instead of treating dataset construction as a sequence of isolated scripts, Med-CRAFT organizes it as a managed workflow in which intermediate artifacts are explicitly represented, stored, inspected, and reused.
This design supports the central requirements defined in Section~\ref{sec:problem}, including traceability, configurability, evidence grounding, reasoning controllability, quality measurability, and reproducibility.

\begin{figure}[htbp]
    \centering
    \includegraphics[width=0.45\textwidth]{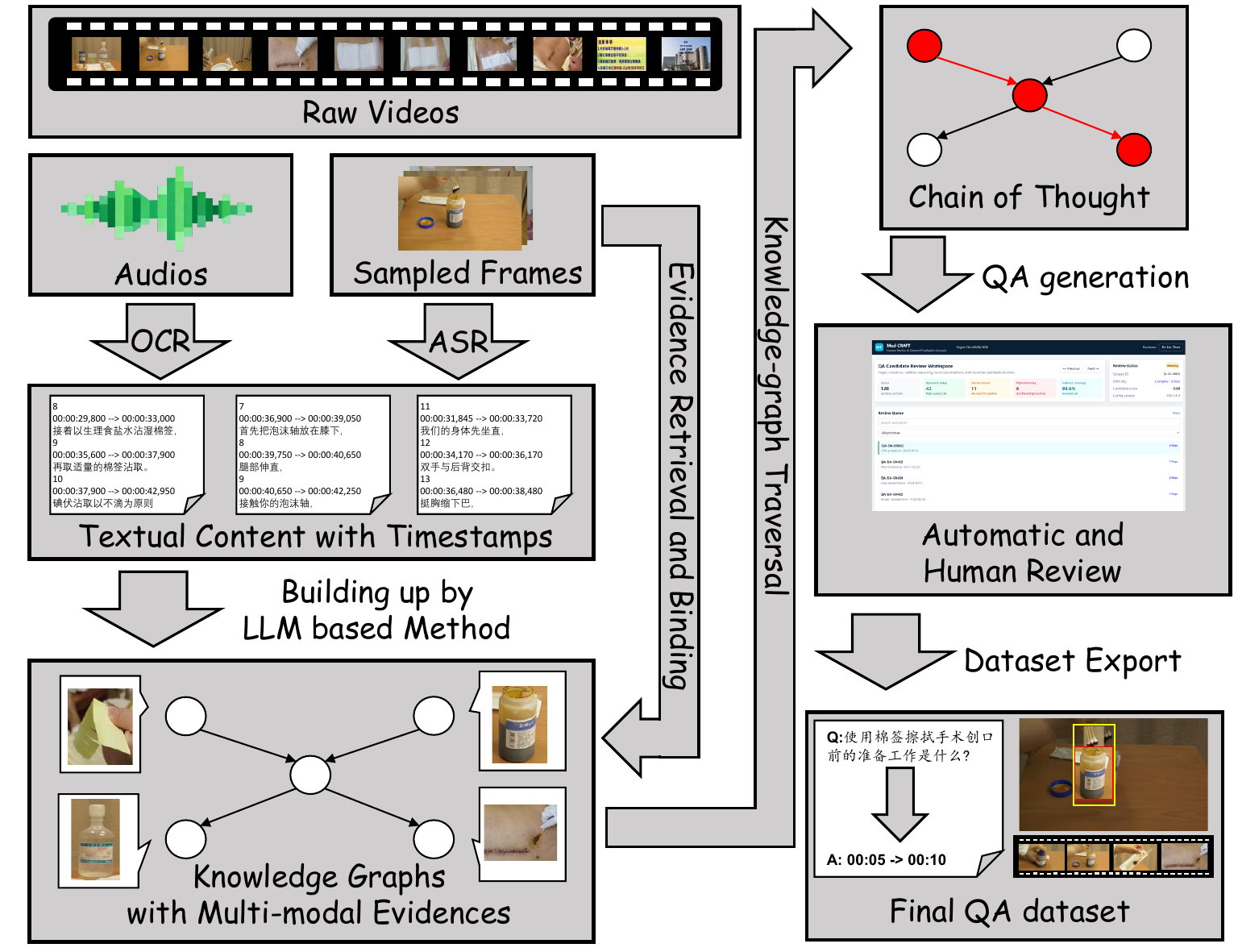}

    \caption{Overview of the Med-CRAFT architecture. The system transforms medical instructional videos into evidence-grounded multimodal QA datasets through multimodal cue extraction, knowledge graph construction, evidence binding, configurable traversal, QA generation, quality filtering, and human review, while maintaining configuration profiles and provenance records throughout the pipeline.}
    \label{fig:system_overview}
\end{figure}

Figure~\ref{fig:system_overview} illustrates the overall architecture of Med-CRAFT.
The architecture consists of six major layers: multimodal video ingestion, medical operation knowledge graph construction, evidence retrieval and binding, graph traversal and reasoning-chain generation, QA generation and quality control, and human-in-the-loop data management.
These layers are supported by a shared storage and logging infrastructure that maintains source videos, extracted modality cues, graph artifacts, evidence indexes, generated samples, configuration files, process logs, and human review records.

\subsection{Multimodal Video Ingestion Layer}

The first layer converts raw instructional videos into temporally aligned multimodal segments.
For each video, Med-CRAFT extracts audio transcripts using automatic speech recognition, on-screen text using optical character recognition, subtitle cues when available, representative frames, shot boundaries, and basic metadata such as duration, resolution, and frame rate.
The extracted signals are aligned by timestamp and stored as video segment objects, which serve as the basic units for subsequent knowledge extraction and evidence retrieval.
This layer also records confidence scores and extraction logs, allowing downstream modules and human reviewers to identify unreliable OCR, ASR, or segmentation results.

\subsection{Medical Operation Knowledge Graph Construction Layer}

The second layer transforms temporally aligned multimodal segments into a medical operation knowledge graph.
Med-CRAFT uses large language model-based information extraction, domain-specific prompts, medical vocabularies, and rule-based normalization to identify entities, events, and relations from OCR text, ASR transcripts, subtitles, and visual descriptions.
The resulting graph represents procedure steps, actions, instruments, anatomical targets, observations, risks, outcomes, and instructional constraints as nodes.
It represents temporal order, causal dependency, tool usage, anatomical location, part-whole structure, and evidential support as typed edges.
Each node and edge is associated with attributes such as normalized labels, source modality, confidence score, candidate evidence segments, and provenance links.
The graph layer therefore acts as an intermediate semantic representation between unstructured multimodal videos and structured QA samples.

\subsection{Evidence Retrieval and Binding Layer}

The third layer enriches the knowledge graph by binding graph elements to concrete evidence units in the source videos.
For each graph node or relation, Med-CRAFT formulates evidence queries based on its label, type, neighboring context, temporal hints, and extracted textual cues.
The system then retrieves candidate video segments by combining lexical matching, temporal proximity, embedding-based semantic retrieval, OCR and ASR alignment, and visual similarity search when frame-level representations are available.
Retrieved candidates are ranked according to relevance, modality consistency, temporal coherence, and confidence scores.
The selected evidence units may include time intervals, key frames, visual regions, OCR spans, ASR spans, subtitle spans, or their combinations.
By binding graph elements to evidence units, Med-CRAFT enables later QA samples to be traced from final answers back to video-level, segment-level, and modality-level evidence.

\subsection{Configurable Graph Traversal and Reasoning-Chain Generation Layer}

The fourth layer generates candidate reasoning chains by traversing the evidence-enriched medical operation knowledge graph.
A traversal configuration specifies the allowed node types, relation types, path length, traversal direction, required evidence coverage, and target question category.
For example, a sequential traversal may follow procedure-step relations, a causal traversal may connect actions to outcomes or risks, and an instrument-centered traversal may connect tools, actions, and anatomical targets.
Each traversed path is converted into a candidate reasoning chain that preserves the graph structure, relation semantics, supporting evidence, and hop count.
The traversal engine rejects paths that are disconnected, weakly evidenced, semantically incoherent, or inconsistent with user-specified constraints.
This layer provides the main mechanism by which Med-CRAFT controls reasoning difficulty and generates diverse QA samples from the same source videos.

\subsection{QA Generation and Quality Control Layer}

The fifth layer verbalizes reasoning chains into natural-language question-answer pairs and packages them with supporting metadata.
Given a reasoning chain, Med-CRAFT generates a question, an answer, an optional explanation, and evidence references using template-based rules, large language model prompting, or a hybrid strategy.
The generation process is constrained by graph semantics, evidence availability, medical terminology, answerability requirements, and user-specified question types.
After generation, each sample is evaluated by automatic quality filters that check duplicate questions, missing evidence, unsupported answers, invalid graph paths, excessive ambiguity, and inconsistent reasoning chains.
Samples that pass automatic filtering can be sent to human reviewers for verification, correction, rejection, or approval.
The final QA sample stores not only the generated textual content, but also the graph path, evidence segments, provenance chain, configuration parameters, quality scores, and review status.

\subsection{Human-in-the-Loop Data Management Layer}

The sixth layer provides a user-facing interface for managing, inspecting, configuring, and validating the dataset construction process.
Through this interface, users can upload videos, configure extraction and traversal parameters, inspect OCR and ASR outputs, visualize knowledge graphs, verify evidence bindings, review generated QA samples, and export finalized datasets.
The interface also provides dashboards for monitoring process-level indicators, such as extraction confidence, graph size, evidence coverage, traversal success rate, filtering rate, review workload, and sample acceptance rate.
These dashboards help users identify processing bottlenecks, compare different configurations, discover systematic errors, and decide where human verification is most needed.
All user operations, configuration changes, review decisions, and exported dataset versions are recorded to support reproducibility and auditability.

\subsection{End-to-End Data Flow}

The end-to-end data flow of Med-CRAFT can be summarized as a sequence of transformations from raw videos to traceable QA samples.
First, raw videos are segmented and converted into multimodal segment objects with aligned textual, audio, visual, and temporal cues.
Second, these segment objects are used to construct medical operation knowledge graphs that encode procedural semantics and candidate evidence links.
Third, graph nodes and edges are bound to concrete temporal, spatial, textual, and visual evidence units retrieved from the original videos.
Fourth, configurable traversal strategies generate reasoning chains from the evidence-enriched graphs.
Fifth, reasoning chains are verbalized into question-answer pairs and filtered according to quality, evidence, and consistency constraints.
Finally, human reviewers inspect selected samples and the system exports finalized datasets together with their provenance records, configurations, quality indicators, and version metadata.

This data flow ensures that every exported sample can be traced backward from the final QA pair to its reasoning chain, graph path, evidence segments, extracted cues, processing logs, and source video.
Conversely, users can also trace forward from a source video segment to the graph elements, reasoning chains, and QA samples derived from it.

\subsection{Design Principles}

The design of Med-CRAFT is guided by four principles.
First, intermediate representations should be explicit, because implicit transformations make generated datasets difficult to inspect and reproduce.
Second, evidence should be treated as a first-class object, because multimodal medical QA samples are meaningful only when answers can be grounded in appropriate textual, visual, and temporal contexts.
Third, generation should be configurable, because different datasets may require different question types, reasoning depths, evidence dependencies, and quality thresholds.
Fourth, automation should be complemented by human verification, because medical data construction involves domain knowledge, safety considerations, and high requirements for semantic correctness.
Together, these principles allow Med-CRAFT to serve not only as a dataset generation pipeline, but also as a provenance-aware and user-controllable data construction information system.

Table~\ref{tab:system_dataflow_notation} summarizes the input-output data structures of the Med-CRAFT workflow and establishes the notation used throughout the subsequent sections.

\begin{table*}[htbp]
\small
\centering
\caption{System-level data flow, intermediate representations, and notation conventions in Med-CRAFT.}
\label{tab:system_dataflow_notation}
\renewcommand{\arraystretch}{1.22}
\setlength{\tabcolsep}{4pt}
\resizebox{\textwidth}{!}{
\begin{tabular}{p{2.6cm}p{2.6cm}p{3.0cm}p{3.0cm}p{4.2cm}p{2.2cm}}
\toprule
\textbf{System Layer / Stage} &
\textbf{Input Object} &
\textbf{Main Transformation} &
\textbf{Output Object} &
\textbf{Representative Data Structure and Symbol} &
\textbf{Cross-Cutting Records} \\
\midrule

Multimodal video ingestion &
Raw medical instructional video collection &
Video parsing, metadata extraction, temporal segmentation, OCR, ASR, subtitle parsing, and frame sampling &
Temporally aligned video segments and multimodal cues &
$\mathcal{V}=\{v_i\}_{i=1}^{N}$, \newline
where $v_i=(id_i,meta_i,R_i,\mathcal{S}_i)$; \newline
$\begin{aligned}
    s_{ij}=& (sid_{ij},t^{start}_{ij},t^{end}_{ij},F_{ij}, \\
            &A_{ij},T^{ocr}_{ij},T^{asr}_{ij},T^{sub}_{ij},\\
            &B_{ij},C_{ij})
\end{aligned}$
&
Configuration profile $\theta^{seg},\theta^{ext}$; extraction logs; source provenance \\

\midrule

Medical operation knowledge graph construction &
Segment-level multimodal cues &
Schema-constrained medical entity, action, instrument, state, relation, and instructional-intent extraction &
Local medical operation knowledge graphs &
$\mathcal{C}=\{C_{ij}\}$; \newline
$G_i=(V_i,E_i,A_i,P_i)$; \newline
$\begin{aligned}
    u=&(uid,type,label,norm,\\
    &attr,conf,ev,prov);
\end{aligned}$ \newline
$\begin{aligned}
    e=&(eid,u_s,u_t,rel,\\
    &attr,conf,ev,prov)
\end{aligned}$ &
Schema version; LLM/model version; prompt identifier; confidence scores \\

\midrule

Knowledge graph consolidation &
Segment-level graphs and terminology candidates &
Entity normalization, coreference resolution, synonym alignment, duplicate merging, and unsupported-element removal &
Video-level consolidated medical operation knowledge graph &
$\mathcal{G}=\{G_i\}_{i=1}^{N}$; \newline
$G_i^{*}=(V_i^{*},E_i^{*},A_i^{*},P_i^{*})$ &
Normalization rules; terminology resources; merge history; graph version \\

\midrule

Evidence retrieval and backward binding &
Graph nodes, edges, source cues, and multimodal indexes &
Textual, visual, temporal, semantic, and graph-context retrieval; relevance ranking; threshold-based evidence attachment &
Evidence-enriched graph &
~\newline
$\begin{aligned}
    \epsilon=&(eid,video,segment,\\
    &interval,modality,region, \\
    &content,score,role);
\end{aligned}$ \newline
$\mathcal{G}^{+}=\mathcal{G}\cup\mathcal{E}$ &
Retrieval configuration $\theta^{ev}$; ranking scores; index version; binding decisions \\

\midrule

Configurable graph traversal and reasoning-chain generation &
Evidence-enriched graph and traversal policy &
Path search, relation-pattern matching, evidence coverage checking, ambiguity filtering, and semantic step ordering &
Candidate reasoning chains &
~\newline
$\begin{aligned}
    \rho_k=&(path_k,steps_k,hop_k,\\
    &pattern_k,ev_k,conf_k);
\end{aligned}$ \newline
$\mathcal{R}=\{\rho_k\}_{k=1}^{K}$ &
Traversal configuration $\theta^{trav}$; path constraints; discarded-path records \\

\midrule

QA generation and automatic quality filtering &
Validated reasoning chains, graph context, and evidence objects &
Question-answer verbalization, rationale generation, answerability validation, consistency checking, and quality scoring &
Candidate QA samples with quality decisions &
~\newline
$\begin{aligned}
    x_k=&(qid,q,a,type,\rho,E,G,P,\\
    &\theta,M,status);
\end{aligned}$ \newline
$\mathcal{Q}=\{x_k\}_{k=1}^{K}$ &
Generation configuration $\theta^{gen}$; filter configuration $\theta^{filter}$; prompts; decoding parameters; quality indicators \\

\midrule

Human-in-the-loop review and dataset finalization &
QA candidates, evidence intervals, graph paths, quality indicators, and provenance records &
Human inspection, correction, approval, rejection, relabeling, and standardized dataset export &
Finalized multimodal medical QA dataset &
$\begin{aligned}
    \mathcal{D}=\{x_k|status(x_k)=\texttt{approved}\};
\end{aligned}$ \newline
$\begin{aligned}
    a_r=&(rid,reviewer,decision,\\
    &comment,time,version)
\end{aligned}$ &
Reviewer actions; sample versions; export version; audit trail \\

\midrule

Pipeline-wide management &
All data objects and transformation outputs &
Configuration control, provenance tracking, quality monitoring, dependency management, backward tracing, and forward impact analysis &
Reproducible and auditable dataset-construction lifecycle &
~\newline
$\begin{aligned}
    \theta=&(\theta^{seg},\theta^{ext},\theta^{kg},\theta^{ev},\\
    &\theta^{trav},\theta^{gen},\theta^{filter},\theta^{review})
\end{aligned}$ \newline
$\begin{aligned}
    p=&(pid,source,operation,\\
    &input,output,actor,time,\\
    &model,parameter,parent);
\end{aligned}$ \newline
$\begin{aligned}
    m=&(mid,target,metric,\\
    &value,method,threshold,\\
    &decision,time)
\end{aligned}$ &
Configuration repository; provenance graph; execution logs; quality dashboards \\

\bottomrule
\end{tabular}
}
\end{table*}

\section{Core Data Models}
\label{sec:data_models}

The effectiveness of Med-CRAFT depends not only on its processing pipeline, but also on the data models used to represent videos, extracted knowledge, evidence, reasoning chains, generated samples, configurations, and provenance records.
In this section, we define the core data models that allow Med-CRAFT to transform unstructured medical instructional videos into traceable and configurable multimodal QA datasets.
These models are designed to satisfy four requirements: preserving multimodal evidence, representing procedural medical semantics, supporting configurable reasoning generation, and maintaining end-to-end provenance.

\subsection{Video and Segment Model}

A medical instructional video is the primary source object in Med-CRAFT.

Each video $v_i$ is represented as a tuple:
\begin{align*}
v_i=(id_i, meta_i, R_i, \mathcal{S}_i),
\end{align*}
where $id_i$ is the unique video identifier, $meta_i$ stores video-level metadata, $R_i$ denotes raw resources, and $\mathcal{S}_i$ is the set of temporally aligned segments derived from the video.
The metadata $meta_i$ may include title, source URL, creator, publication date, medical specialty, language, duration, resolution, frame rate, and license information.
The raw resources $R_i$ include the original video file, audio stream, subtitle file if available, extracted frames, and any user-provided auxiliary documents.

The segment set $\mathcal{S}_i=\{s_{i1},s_{i2},\ldots,s_{iM_i}\}$ provides the basic temporal units for extraction, evidence retrieval, and answer grounding.
Each segment $s_{ij}$ is represented as:
\begin{align*}
s_{ij}=(sid_{ij}, t^{start}_{ij}, t^{end}_{ij}, F_{ij}, A_{ij}, T^{ocr}_{ij}, T^{asr}_{ij}, T^{sub}_{ij}, B_{ij}, C_{ij}),
\end{align*}
where $sid_{ij}$ is the segment identifier, $t^{start}_{ij}$ and $t^{end}_{ij}$ define the temporal interval, $F_{ij}$ denotes sampled frames, $A_{ij}$ denotes the corresponding audio clip, and $T^{ocr}_{ij}$, $T^{asr}_{ij}$, and $T^{sub}_{ij}$ denote OCR text, ASR transcript, and subtitle text, respectively.

The field $B_{ij}$ stores optional spatial boundaries, such as text bounding boxes, detected object regions, or manually marked regions of interest.
The field $C_{ij}$ stores modality-specific confidence scores and extraction diagnostics, such as OCR confidence, ASR confidence, segmentation confidence, and frame selection quality.
This segment-level representation enables Med-CRAFT to align medical semantics with concrete temporal and spatial evidence.

\subsection{Medical Operation Knowledge Graph Model}

The medical operation knowledge graph is the central semantic representation in Med-CRAFT.

For each video $v_i$, Med-CRAFT constructs a graph:
\begin{align*}
G_i=(V_i,E_i,A_i,P_i),
\end{align*}
where $V_i$ is the set of nodes, $E_i$ is the set of typed edges, $A_i$ is the set of node and edge attributes, and $P_i$ is the set of provenance records associated with graph elements.

A node $u \in V_i$ is represented as:
\begin{align*}
u=(uid, type, label, norm, attr, conf, ev, prov),
\end{align*}
where $uid$ is the node identifier, $type$ denotes the node type, $label$ is the extracted surface form, $norm$ is the normalized medical or procedural concept, $attr$ stores additional attributes, $conf$ is the extraction confidence score, $ev$ stores evidence links, and $prov$ stores provenance information.
Node types include \textit{ProcedureStep}, \textit{Action}, \textit{Instrument}, \textit{AnatomicalStructure}, \textit{Observation}, \textit{Risk}, \textit{Outcome}, \textit{Instruction}, and \textit{Constraint}.

An edge $e \in E_i$ is represented as:
\begin{align*}
e=(eid, u_s, u_t, rel, attr, conf, ev, prov),    
\end{align*}
where $eid$ is the edge identifier, $u_s$ and $u_t$ are the source and target nodes, $rel$ is the relation type, $attr$ stores relation attributes, $conf$ is the relation extraction confidence, $ev$ stores supporting evidence links, and $prov$ stores the derivation history of the edge.
Relation types include \textit{precedes}, \textit{follows}, \textit{uses}, \textit{acts-on}, \textit{located-in}, \textit{causes}, \textit{prevents}, \textit{indicates}, \textit{part-of}, \textit{requires}, and \textit{evidenced-by}.

This graph model allows Med-CRAFT to represent medical instructional content as a structured set of entities, events, relations, and evidence-supported procedural dependencies.

\subsection{Evidence Model}

Evidence is treated as a first-class object in Med-CRAFT because every generated QA sample should be grounded in identifiable video content.
An evidence unit $\epsilon$ is represented as:
\begin{align*}
\epsilon=(eid, video, segment, interval, modality, region, content, score, role),    
\end{align*}
where $eid$ is the evidence identifier, $video$ identifies the source video, $segment$ identifies the segment, $interval$ specifies the temporal range, $modality$ indicates the evidence type, $region$ stores spatial information if available, $content$ stores textual or visual descriptions, $score$ denotes the relevance or confidence score, and $role$ specifies how the evidence supports a graph element or QA sample.
The modality field may take values such as \textit{visual}, \textit{ocr}, \textit{asr}, \textit{subtitle}, \textit{metadata}, or \textit{multimodal}.
The role field may indicate whether the evidence supports an entity mention, a relation, an answer, a reasoning step, a temporal boundary, or a visual grounding requirement.
For example, an \textit{Instrument} node may be supported by an OCR span naming the tool, an ASR span explaining its usage, and a visual region showing the tool being manipulated.
By representing evidence explicitly, Med-CRAFT can distinguish text-only answerability from visually grounded answerability and can measure the degree to which each sample depends on multimodal evidence.

\subsection{Reasoning Chain Model}

A reasoning chain represents the structured path used to derive a question-answer sample from the evidence-enriched knowledge graph.
Given a graph $G_i$, a reasoning chain $\rho_k$ for sample $x_k$ is defined as:
\begin{align*}
\rho_k=(path_k, steps_k, hop_k, pattern_k, ev_k, conf_k),
\end{align*}
where $path_k$ is the traversed graph path or subgraph, $steps_k$ is the ordered list of reasoning statements, $hop_k$ is the number of reasoning hops, $pattern_k$ describes the traversal pattern, $ev_k$ stores supporting evidence units, and $conf_k$ denotes the estimated reliability of the chain.
The traversal pattern may be \textit{sequential}, \textit{causal}, \textit{instrument-action-object}, \textit{anatomy-centered}, \textit{risk-prevention}, or \textit{cross-modal-evidence}.
Each reasoning step should be aligned with at least one graph edge or evidence unit, so that the chain remains faithful to the underlying data rather than becoming a free-form explanation.
This model allows Med-CRAFT to control the reasoning complexity of generated QA samples and to analyze model performance by hop count, relation pattern, and evidence dependency.

\subsection{Question-Answer Sample Model}

The final dataset instance in Med-CRAFT is represented as a rich QA sample rather than a simple question-answer pair.
A QA sample $x_k$ is defined as:
\begin{align*}
x_k=(qid, q, a, type, \rho, E, G, P, \theta, M, status),    
\end{align*}
where $qid$ is the question identifier, $q$ is the natural-language question, $a$ is the answer, $type$ denotes the question category, $\rho$ is the reasoning chain, $E$ is the set of evidence units, $G$ is the associated graph path or subgraph, $P$ is the provenance record, $\theta$ is the generation configuration, $M$ is the set of quality metrics, and $status$ records the review state.
Question categories may include recognition, temporal localization, procedural ordering, tool usage, anatomical grounding, causal reasoning, risk prevention, and cross-modal multi-hop reasoning.
The review status may take values such as \textit{generated}, \textit{auto-filtered}, \textit{pending-review}, \textit{accepted}, \textit{revised}, or \textit{rejected}.
This representation allows downstream users to filter samples by question type, reasoning depth, evidence modality, review status, confidence score, or provenance completeness.

\subsection{Configuration Model}

Configurability is implemented through an explicit configuration model that records how each dataset construction run is parameterized.
A configuration $\theta$ is represented as:
\begin{align*}
\theta=(\theta^{seg}, \theta^{ext}, \theta^{kg}, \theta^{ev}, \theta^{trav}, \theta^{gen}, \theta^{filter}, \theta^{review}),
\end{align*}
where $\theta^{seg}$ controls video segmentation, $\theta^{ext}$ controls OCR, ASR, and modality extraction, $\theta^{kg}$ controls knowledge graph construction, $\theta^{ev}$ controls evidence retrieval and ranking, $\theta^{trav}$ controls graph traversal, $\theta^{gen}$ controls QA generation, $\theta^{filter}$ controls quality filtering, and $\theta^{review}$ controls human review policies.

For example, $\theta^{trav}$ may specify the maximum hop count, allowed relation types, required evidence coverage, target question distribution, and traversal templates.
Similarly, $\theta^{filter}$ may specify thresholds for evidence confidence, answerability, duplicate similarity, graph-path validity, and medical terminology consistency.
By storing $\theta$ with each dataset version and each generated sample, Med-CRAFT makes it possible to reproduce, compare, and audit different dataset construction runs.

\subsection{Provenance Model}

The provenance model records the derivation history of each artifact produced by Med-CRAFT.
A provenance record $p$ is represented as:
\begin{align*}
p=(pid, source, operation, input, output,   
\end{align*}
\vspace{-0.5cm}
\begin{align*}
actor, time, model, parameter, parent), 
\end{align*}
where $pid$ is the provenance identifier, $source$ identifies the original resource, $operation$ describes the transformation, $input$ and $output$ identify consumed and produced artifacts, $actor$ denotes the system module or human reviewer, $time$ records the timestamp, $model$ records the algorithm or model version, $parameter$ records the relevant configuration, and $parent$ links to previous provenance records.
A QA sample may therefore have a provenance chain linking it to QA generation, reasoning-chain construction, graph traversal, evidence binding, knowledge extraction, modality extraction, segmentation, and the original video.
This chain allows users to answer questions such as which video segment supports an answer, which graph path generated the question, which model extracted a relation, and which reviewer approved the final sample.
The provenance model is essential for debugging, reproducibility, accountability, and trustworthy dataset release.

\subsection{Quality Indicator Model}

Med-CRAFT records quality indicators at both process level and sample level.
Process-level indicators describe the behavior and reliability of construction modules.
These indicators include OCR confidence, ASR confidence, segment coverage, graph node count, graph edge count, graph density, evidence coverage, traversal success rate, invalid path rate, duplicate generation rate, filtering rate, review workload, and acceptance rate.
Sample-level indicators describe the quality of individual QA samples.
These indicators include semantic consistency, medical validity, answerability, evidence relevance, temporal localization accuracy, reasoning-chain faithfulness, visual-dependency level, language fluency, and reviewer agreement.
A quality indicator record $m$ is represented as:
\begin{align*}
m=(mid, target, metric, value, method, threshold, decision, time),    
\end{align*}
where $target$ identifies the process, artifact, or sample being evaluated, $metric$ denotes the metric name, $value$ stores the measured value, $method$ records whether the value is computed automatically or assigned by human reviewers, $threshold$ stores the decision threshold, and $decision$ records whether the target passes the corresponding quality criterion.
By integrating quality indicators with provenance and configuration records, Med-CRAFT enables systematic analysis of how construction settings affect dataset quality.

\subsection{Relationships Among Data Models}

The data models in Med-CRAFT are connected through explicit identifiers and provenance links.
A video contains segments, segments support graph nodes and edges, graph elements are linked to evidence units, evidence-enriched graph paths form reasoning chains, and reasoning chains are verbalized into QA samples.
Configurations determine how each transformation is performed, provenance records describe how each artifact is derived, and quality indicators evaluate the reliability of both intermediate artifacts and final samples.
This relational structure enables both backward tracing from a QA sample to its source video and forward tracing from a video segment to all derived graph elements, reasoning chains, and QA samples.
The resulting data model design provides the structural foundation for the construction pipeline described in the next section.

\section{Med-CRAFT Pipeline}
\label{sec:pipeline}

The Med-CRAFT pipeline operationalizes the abstract data models described above into an executable workflow for constructing multimodal medical question-answering datasets from instructional videos, following the broader principle that data-centric AI systems should make data construction processes measurable, iterative, and controllable \cite{ng2021data,mazumder2022dataperf,zha2023data}.
Instead of treating dataset construction as a single end-to-end generation task, Med-CRAFT decomposes it into a sequence of auditable transformations over videos, segments, cues, graphs, evidence objects, reasoning chains, and QA samples, which is consistent with provenance-aware workflow management in scientific data systems \cite{simmhan2005survey,herschel2017survey,stoyanovich2022responsible}.
This decomposition is essential for data-intensive AI applications because each intermediate representation can be inspected, configured, measured, and revised before it affects the final dataset \cite{ratner2017snorkel,ratner2020snorkel}.

Formally, the pipeline can be summarized as follows:
\begin{align*}
\mathcal{V}
\rightarrow
\mathcal{S}
\rightarrow
\mathcal{C}
\rightarrow
\mathcal{G}
\rightarrow
\mathcal{G}^{+}
\rightarrow
\mathcal{R}
\rightarrow
\mathcal{Q}
\rightarrow
\mathcal{D},    
\end{align*}
where $\mathcal{V}$ denotes the input video collection, $\mathcal{S}$ denotes temporally segmented video units, $\mathcal{C}$ denotes extracted multimodal cues, $\mathcal{G}$ denotes initial medical operation knowledge graphs, $\mathcal{G}^{+}$ denotes evidence-enriched graphs, $\mathcal{R}$ denotes reasoning chains, $\mathcal{Q}$ denotes candidate QA samples, and $\mathcal{D}$ denotes the finalized dataset.
Table~\ref{tab:system_dataflow_notation} also shown that the Med-CRAFT pipeline transforms raw videos into finalized QA samples through a sequence of explicit intermediate representations.

\subsection{Video Preprocessing}

The first stage takes raw medical instructional videos as input and converts them into temporally organized and machine-processable units, which is a common prerequisite for instructional video understanding and medical video question answering \cite{gupta2022medvidqa,gupta2023medvidqa,m3med2025}.
For each video, Med-CRAFT extracts metadata such as title, source, language, duration, publication information, and available textual descriptions.
The system then performs temporal segmentation according to configurable rules, including fixed-length windows, subtitle boundaries, shot changes, silence intervals, or detected procedural transitions, since answer localization in medical videos often depends on accurate temporal units rather than whole-video representations \cite{gupta2022medvidqa,gupta2024trecmedvidqa}.

Each resulting segment is assigned a unique identifier and linked to its original video, temporal interval, frame sequence, audio track, and preprocessing configuration.
This design allows later QA samples to be traced back not only to a source video but also to the exact temporal unit from which their evidence was derived.

\subsection{Multimodal Cue Extraction}

After segmentation, Med-CRAFT extracts multimodal cues from each video segment using OCR, ASR, subtitle parsing, frame sampling, and visual feature encoding, reflecting the common observation that medical instructional video understanding requires both textual and visual evidence \cite{gupta2023medvidqa,m3med2025}.
Textual cues include recognized on-screen text, transcribed speech, aligned subtitles, video descriptions, and automatically detected medical terms, which are especially important because many health-related video QA datasets provide answer-bearing information through narration or subtitles \cite{gupta2022medvidqa,gupta2023medvidqa}.
Visual cues include sampled frames, object regions, hand-object interactions, tool appearances, anatomical references, and local visual embeddings, since recent medical instructional video benchmarks show that complex questions often require grounding in visual demonstrations rather than relying only on transcripts \cite{m3med2025}.
Temporal cues include segment boundaries, action durations, ordering relations, and overlaps between spoken explanations and visual demonstrations.

All extracted cues are stored with confidence scores, extraction methods, timestamps, and links to the corresponding segment records.
By preserving heterogeneous cues before knowledge extraction, the system avoids prematurely collapsing multimodal information into a single textual representation, which is important for evaluating whether models truly integrate visual, textual, and temporal evidence \cite{m3med2025,hong2023bestmvqa}.

\subsection{Medical Operation Knowledge Extraction}

The third stage transforms multimodal cues into structured medical operation knowledge using LLM-based extraction under constrained schemas, following recent work that uses large language models to support information extraction and knowledge graph construction from unstructured data \cite{tan2024llmannotation,liu2025llmdatasets,zhu2024mmkg}.

Given the cue set of a segment, the extractor identifies medical entities, procedural actions, instruments, anatomical objects, states, conditions, temporal relations, causal relations, and instructional intents, which are represented as graph elements to support later reasoning and question generation \cite{yih2016semantic,bordes2015large,liu2025kbqg}.
The extraction prompt is parameterized by a domain schema that specifies valid node types, relation types, attribute fields, and confidence requirements, because schema-guided extraction can reduce unconstrained generation and improve structural consistency \cite{tan2024llmannotation,zhu2024mmkg}.
The extractor is instructed to abstain from creating graph elements when the evidence is insufficient or when the medical meaning cannot be reliably inferred from the available cues, which is necessary because LLM-generated annotations may otherwise introduce unsupported or hallucinated content \cite{tan2024llmannotation,liu2025llmdatasets}.

For example, an action node may be required to contain an action label, involved object, temporal interval, required instrument, and supporting cue identifiers.
The extractor is instructed to abstain from creating graph elements when the evidence is insufficient or when the medical meaning cannot be reliably inferred from the available cues.
Each extracted node and edge is associated with its source cues so that later reviewers can inspect why a particular graph element was created.

\subsection{Knowledge Graph Consolidation}

Since different segments of the same video may mention the same medical object, action, or state using different expressions, Med-CRAFT performs graph consolidation after local extraction, following common knowledge graph construction practice that requires entity normalization, relation validation, and knowledge fusion \cite{chen2023mmkgsurvey,zhu2024mmkg}.
This stage normalizes entity names, merges duplicate nodes, resolves coreference, aligns synonymous medical terms, and removes low-confidence or unsupported graph elements, which helps improve the semantic coherence of the video-level graph \cite{chen2023mmkgsurvey}.

The consolidation process combines lexical similarity, embedding similarity, terminology resources, temporal proximity, and LLM-based verification.
When two candidate nodes are merged, the system preserves their original identifiers as provenance links rather than deleting their construction history.
The output of this stage is a video-level medical operation knowledge graph that represents procedural structure, semantic dependencies, and temporal progression.

\subsection{Evidence Retrieval and Backward Binding}

After constructing the initial knowledge graph, Med-CRAFT performs backward evidence binding to associate graph elements with concrete video evidence, aligning with the evidence-grounded evaluation requirement in medical video QA and multimodal reasoning benchmarks \cite{gupta2023medvidqa,gupta2024trecmedvidqa,m3med2025}.

For each node or edge, the system generates a retrieval query from its label, attributes, neighboring graph context, and original source cues.
Candidate evidence segments are retrieved from textual indexes, visual indexes, temporal indexes, and multimodal embedding indexes, following the general practice of combining sparse, dense, and multimodal retrieval signals for evidence grounding \cite{karpukhin2020dense,radford2021clip,lei2021clipbert}.
The relevance score between a graph element $z$ and an evidence object $\epsilon$ is computed as:
\begin{align*}
Score(z,\epsilon)=
\lambda_1 Sim_{text}(z,\epsilon)
+\lambda_2 Sim_{sem}(z,\epsilon)
\end{align*}
\vspace{-0.5cm}
\begin{align*}
+\lambda_3 Sim_{vis}(z,\epsilon)
+\lambda_4 Temp(z,\epsilon)
+\lambda_5 Conf(\epsilon),
\end{align*}
where $Sim_{text}$ measures lexical overlap, $Sim_{sem}$ measures semantic similarity, $Sim_{vis}$ measures visual relevance, $Temp$ measures temporal compatibility, and $Conf$ denotes extraction confidence.

The weights $\lambda_1,\ldots,\lambda_5$ are configurable so that different dataset construction settings can emphasize textual grounding, visual grounding, temporal precision, or conservative confidence filtering.
Only evidence objects that satisfy predefined relevance and temporal-overlap thresholds are attached to the corresponding graph elements.
The resulting evidence-enriched graph $\mathcal{G}^{+}$ provides a structured bridge between symbolic reasoning paths and observable multimodal video content, which is crucial for distinguishing grounded reasoning from answer generation based only on textual correlations \cite{m3med2025,hong2023bestmvqa}.

\begin{figure}[htbp]
    \centering
    \includegraphics[width=0.45\textwidth]{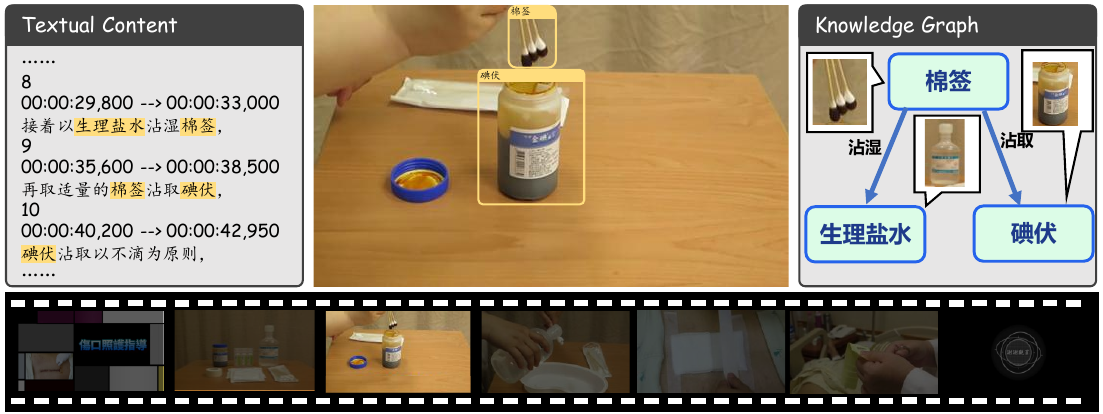}

    \caption{Evidence Binding Example}
    \label{fig:evidence_binding_example}
\end{figure}

Figure~\ref{fig:evidence_binding_example} gives an illustrative example of backward evidence binding, where graph nodes and relations are linked to textual, visual, and temporal evidence intervals in the source video.

\subsection{Configurable Graph Traversal}

The next stage traverses the evidence-enriched graph to generate candidate reasoning chains with controllable complexity, inspired by knowledge-graph-based multi-hop reasoning and question generation methods \cite{yih2016semantic,bordes2015large,liu2025kbqg}.
A traversal configuration specifies the allowed starting node types, ending node types, relation patterns, maximum hop length, minimum evidence coverage, and target reasoning categories, enabling explicit control over reasoning complexity as required by multi-hop QA benchmarks \cite{m3med2025,liu2025kbqg}.

For example, a one-hop chain may connect an action to its required instrument, whereas a multi-hop chain may connect a symptom, a diagnostic action, an observed anatomical region, and a final interpretation.
Med-CRAFT supports several traversal patterns, including action-object paths, action-instrument paths, temporal-before-after paths, condition-action paths, and evidence-composition paths.

Each candidate path is converted into a reasoning chain by ordering its graph elements, collecting their attached evidence, and generating an explicit step-by-step semantic description.
Chains that lack sufficient evidence, contain unsupported medical relations, or exceed the configured ambiguity threshold are discarded.
This traversal mechanism enables Med-CRAFT to generate QA samples with different difficulty levels while keeping their reasoning structures transparent, which differs from direct LLM-based QA generation where the underlying reasoning path is often implicit \cite{tan2024llmannotation,liu2025llmdatasets}.

\begin{figure}[htbp]
    \centering
    \includegraphics[width=0.45\textwidth]{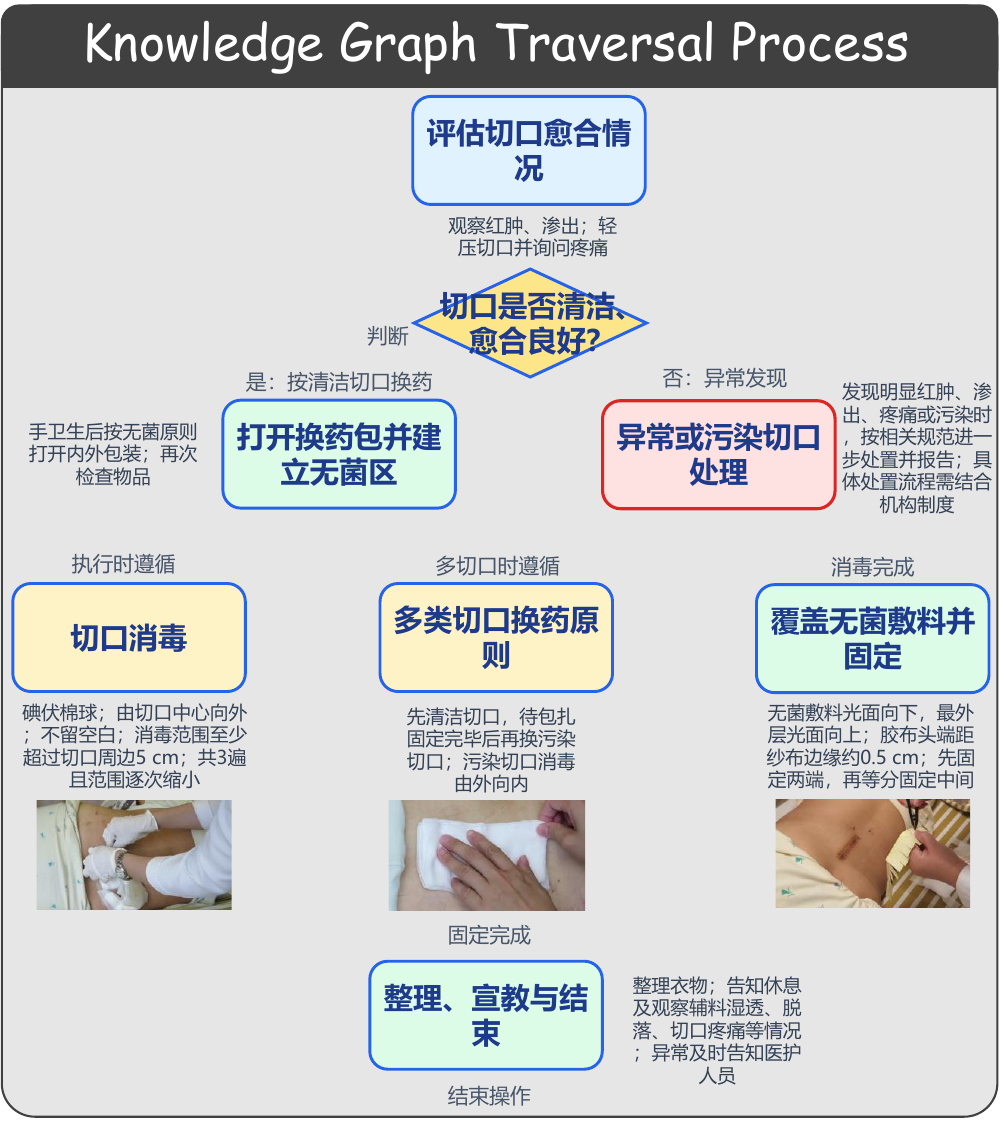}

    \caption{From Graph Traversal to QA Generation Example}
    \label{fig:traversal_to_qa}
\end{figure}

Figure~\ref{fig:traversal_to_qa} illustrates how a configured graph path is converted into a reasoning chain and then into an evidence-grounded QA sample.

\subsection{Reasoning-Chain-Based QA Generation}

Given a validated reasoning chain, Med-CRAFT generates a natural-language question, a reference answer, and supporting evidence annotations, following the line of work that uses structured knowledge to guide controllable and answerable question generation \cite{serban2016generating,liu2025kbqg,graphguided2024procedural}.
The generator receives the reasoning chain, graph context, evidence snippets, target question type, expected answer granularity, and linguistic constraints as input.

Question types may include factual identification, temporal localization, procedural reasoning, causal explanation, condition-based decision, and cross-modal evidence integration, reflecting the ability requirements emphasized by medical video QA and multimodal multi-hop reasoning benchmarks \cite{gupta2023medvidqa,m3med2025}.
The generated answer is required to be derivable from the reasoning chain and verifiable against the bound evidence intervals, following the principle that benchmark instances should support answer verification and error analysis \cite{gupta2024trecmedvidqa,m3med2025}.
For complex samples, the generator also produces an explicit rationale that maps each reasoning step to one or more evidence objects.

The system stores not only the final question and answer but also the underlying path, intermediate reasoning steps, evidence identifiers, generation prompt, model version, and decoding parameters.
This design makes it possible to distinguish errors caused by graph construction, evidence retrieval, traversal configuration, or language generation.

\subsection{Automatic Quality Filtering}

Before human review, Med-CRAFT applies automatic quality filters to remove low-quality, inconsistent, or weakly grounded QA candidates, following data-centric AI practices that emphasize systematic data validation before model evaluation \cite{ng2021data,mazumder2022dataperf,zha2023data}.

The filtering module evaluates each candidate along multiple dimensions, including question clarity, answer consistency, medical plausibility, evidence relevance, temporal localization accuracy, and reasoning-chain faithfulness, since medical QA datasets require both linguistic validity and domain-specific reliability \cite{lau2018vqarad,he2020pathvqa,liu2021slake}.
A candidate is rejected if its answer cannot be recovered from the evidence, if its rationale contradicts the graph, or if its evidence interval does not contain the required visual or textual cues.
For samples generated from multi-hop chains, the system additionally checks whether all intermediate reasoning steps are supported rather than only the final answer.

The automatic filters are not intended to replace expert judgment but to reduce the review burden by prioritizing candidates with higher structural and evidential reliability, which is consistent with human-in-the-loop data curation practices \cite{ratner2017snorkel,stoyanovich2022responsible}.

\subsection{Human Review and Dataset Finalization}

The remaining candidates are presented to human reviewers through the Med-CRAFT user interface, because human verification remains essential for medical plausibility, safety-sensitive interpretation, and final dataset reliability \cite{lau2018vqarad,he2020pathvqa,liu2021slake}.

For each sample, reviewers can inspect the question, answer, reasoning chain, graph path, evidence intervals, video frames, extracted cues, and provenance records, following the broader requirement that high-stakes datasets should support transparency, accountability, and reviewability \cite{stoyanovich2022responsible,herschel2017survey}.
Reviewers may approve a sample, reject it, edit its wording, adjust its evidence interval, correct its graph linkage, or assign additional quality labels.
All reviewer actions are logged as provenance events and linked to the corresponding sample version.

After review, approved samples are exported into standardized dataset formats containing questions, answers, evidence intervals, reasoning chains, metadata, and evaluation labels, enabling both benchmark evaluation and reproducible dataset reuse \cite{gupta2024trecmedvidqa,m3med2025}.
This finalization stage ensures that the released dataset remains both machine-readable for benchmark evaluation and human-interpretable for auditing and reuse.

\subsection{Pipeline-Level Provenance and Reproducibility}

A central feature of Med-CRAFT is that every pipeline stage produces explicit provenance records, following established research on data provenance for transparency, reproducibility, and workflow auditing \cite{simmhan2005survey,herschel2017survey}.
These records include input identifiers, output identifiers, operation names, software versions, model versions, configuration parameters, execution time, confidence scores, and parent-child dependencies.

Because each QA sample is connected to its graph path, evidence objects, intermediate cues, source segments, and original video, Med-CRAFT supports backward tracing from dataset instances to raw data, which is important for error diagnosis and reproducible experimentation \cite{simmhan2005survey,herschel2017survey}.

Conversely, the system also supports forward tracing from a video segment to all graph elements, reasoning chains, and QA samples derived from it.
This bidirectional traceability is useful for error correction, dataset maintenance, copyright auditing, expert review, and reproducible experimentation, and it directly addresses the lack of traceability often observed in ad hoc dataset construction workflows \cite{herschel2017survey,stoyanovich2022responsible}.

When a configuration is changed, Med-CRAFT can identify which downstream artifacts should be regenerated and which artifacts remain valid.
Therefore, the pipeline is not merely a data generation procedure but a reproducible information system for managing the lifecycle of multimodal medical QA dataset construction \cite{simmhan2005survey,herschel2017survey,stoyanovich2022responsible}.

\section{Implementation}
\label{sec:implementation}

\begin{figure}[htbp]
    \centering
    \includegraphics[width=0.45\textwidth]{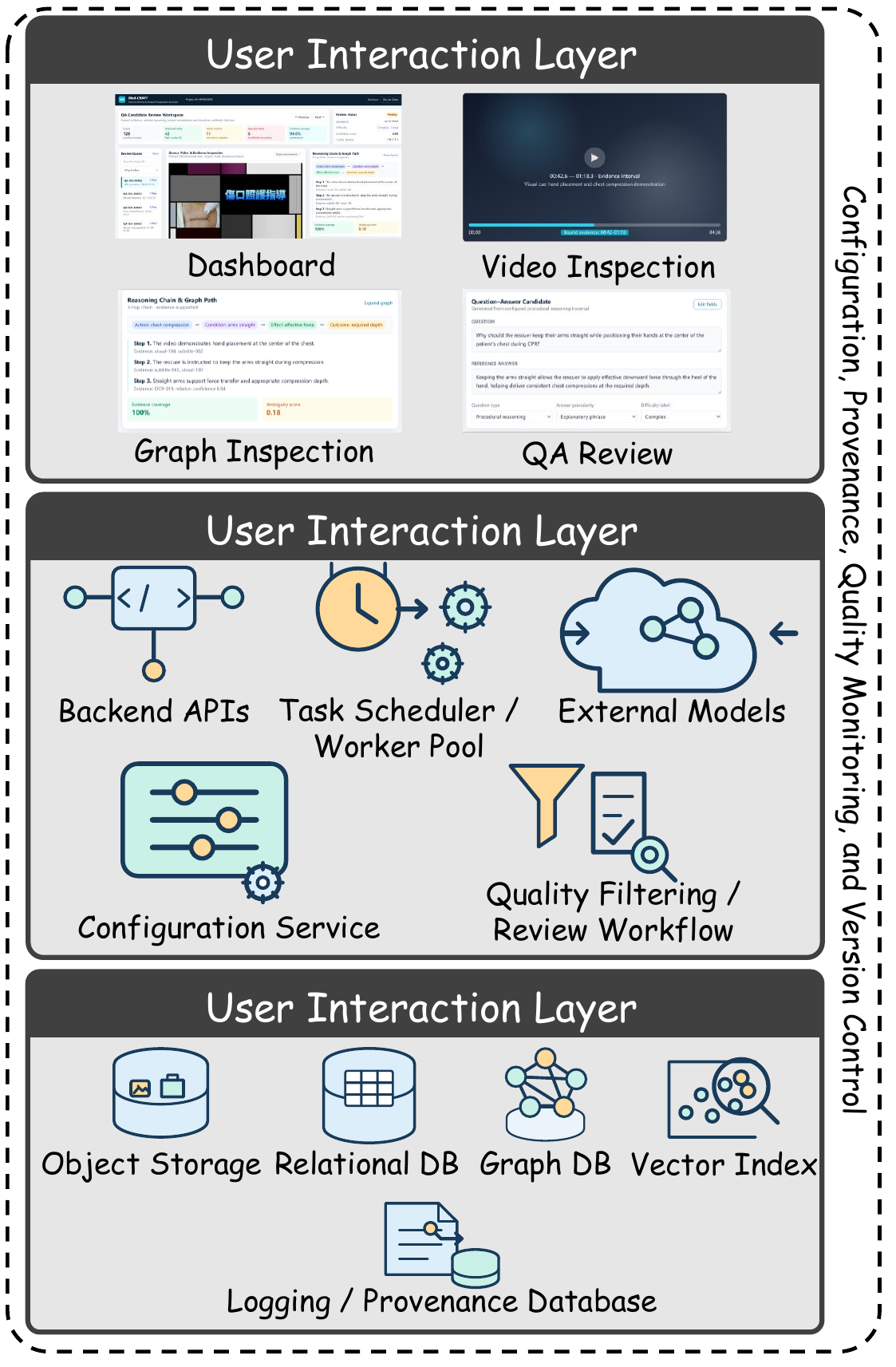}
    \caption{Implementation Architecture / Deployment Diagram}
    \label{fig:implementation_architecture}
\end{figure}

Figure~\ref{fig:implementation_architecture} presents the implementation architecture of Med-CRAFT, including frontend interfaces, backend services, asynchronous workers, heterogeneous storage components, indexes, and external model services.

This section describes the implementation of Med-CRAFT as a modular information system that supports scalable processing, configurable execution, provenance-aware data management, and human-in-the-loop dataset construction.
While the pipeline defines how raw videos are transformed into QA samples, the implementation specifies how these transformations are executed, stored, monitored, versioned, and exposed to users.
The system follows a service-oriented architecture consisting of a backend service layer, a task execution layer, a multimodal storage layer, an indexing layer, a provenance and logging layer, and a web-based user interface.
This architectural design allows different extraction models, graph construction strategies, retrieval methods, and review policies to be replaced without changing the entire system.

\subsection{Backend Architecture}

The backend of Med-CRAFT is implemented as a set of loosely coupled services organized around videos, segments, graphs, evidence objects, reasoning chains, QA samples, configurations, and review records.

Each service provides a unified API for creating, querying, updating, validating, and exporting its corresponding data objects.
The video service manages raw video files, metadata, temporal segments, sampled frames, and links to extracted multimodal cues.
The graph service manages medical operation knowledge graphs, including node creation, edge creation, graph consolidation, evidence attachment, and graph versioning.
The dataset service manages reasoning chains, QA candidates, quality scores, review states, and finalized dataset exports.
The configuration service stores reusable execution profiles so that users can rerun the same construction workflow under identical or modified settings.

\subsection{Task Execution Layer}

Med-CRAFT uses an asynchronous task execution layer to handle computationally expensive operations such as OCR, ASR, frame encoding, LLM-based extraction, graph consolidation, evidence retrieval, and QA generation.

Each task is represented as an executable unit with explicit inputs, outputs, parameters, status, retry policy, and dependency constraints.
The task scheduler constructs a dependency graph among tasks and executes downstream tasks only after their required upstream artifacts have been successfully generated.

For example, QA generation tasks cannot be scheduled until the corresponding reasoning chains and evidence-enriched graph elements are available.
Failed tasks are marked with error types, diagnostic messages, input identifiers, and execution logs so that users can inspect and selectively rerun them.

This execution model improves robustness because a failure in one video, segment, or graph component does not necessarily invalidate the entire dataset construction process.

\subsection{Multimodal Storage Design}

The storage layer is designed to manage heterogeneous data with different access patterns, including raw videos, frames, transcripts, graph structures, embedding vectors, evidence records, and dataset samples.

Large binary objects such as videos, audio files, sampled frames, and visual crops are stored in object storage with stable resource identifiers.
Structured metadata such as segment records, extraction results, configuration profiles, quality indicators, and review states are stored in a relational database.
Medical operation knowledge graphs are stored in a graph database or a graph-compatible relational schema to support efficient path queries and neighborhood inspection.
Embedding vectors for textual, visual, and multimodal retrieval are stored in a vector index that supports approximate nearest-neighbor search.

These storage components are connected through persistent identifiers so that a QA sample can be resolved into its graph path, evidence intervals, source cues, and raw video resources.

\subsection{Indexing and Retrieval Infrastructure}

To support backward evidence binding, Med-CRAFT builds multiple indexes over textual, visual, temporal, and graph-based representations.

The textual index stores OCR tokens, ASR transcripts, subtitles, detected medical terms, and normalized entity mentions.
The visual index stores frame-level embeddings, region-level embeddings, object detection results, and visual descriptions generated by multimodal models.
The temporal index stores segment boundaries, action intervals, subtitle alignment, and overlaps among visual, textual, and audio cues.
The graph index stores node types, relation types, neighborhood structures, and frequently used traversal patterns.

During evidence retrieval, these indexes are queried jointly and their results are fused by configurable scoring functions.
This infrastructure enables Med-CRAFT to retrieve not only semantically similar evidence but also temporally precise and visually grounded evidence.

\subsection{Configuration Management}

Configuration management is implemented as a first-class component rather than as a set of informal script arguments.

A configuration profile specifies segmentation rules, extraction models, prompt templates, graph schemas, retrieval weights, traversal constraints, quality thresholds, and review policies.
Each profile is assigned a unique identifier and version number so that all artifacts generated under that profile can be reproduced or compared with artifacts generated under another profile.
When a user modifies a configuration, the system records the difference between the old and new profiles and estimates the downstream artifacts affected by the change.

For example, changing the maximum traversal hop length affects reasoning-chain generation and QA generation but does not require rerunning OCR or ASR.
This mechanism supports controlled experimentation because researchers can attribute dataset differences to explicit configuration changes.

\subsection{Provenance and Logging Mechanism}

Med-CRAFT implements provenance tracking at the level of tasks, data objects, model calls, user actions, and dataset exports.

For every generated artifact, the system records the operation that produced it, the input artifacts used, the configuration applied, the model version invoked, and the timestamp of execution.
Model calls to LLMs and multimodal models are logged with prompt identifiers, input hashes, output hashes, decoding parameters, and post-processing decisions.
User actions such as approval, rejection, correction, relabeling, and evidence adjustment are also stored as provenance events.
The logging subsystem maintains both human-readable logs for debugging and structured logs for metric aggregation and process analysis.
Together, these records allow users to answer when, why, how, and by whom a particular dataset instance was created or modified.

\subsection{User Interface}

Figure~\ref{fig:implementation_architecture} also shows representative user interface panels of Med-CRAFT, including the dashboard, video inspection panel, graph inspection panel, and QA review panel.

The Med-CRAFT user interface is designed to make the dataset construction process observable, controllable, and reviewable.

The dashboard provides an overview of uploaded videos, processing progress, task failures, extraction statistics, evidence coverage, QA generation counts, and review status.
The video inspection panel allows users to view video segments together with OCR results, ASR transcripts, subtitles, detected entities, sampled frames, and candidate evidence intervals.
The graph inspection panel visualizes medical operation knowledge graphs and supports filtering by node type, relation type, confidence score, evidence availability, and review state.
The QA review panel presents each candidate sample with its question, answer, reasoning chain, evidence clips, graph path, quality indicators, and provenance summary.

Reviewers can directly approve, reject, edit, relabel, or annotate a sample, and every action is immediately reflected in the dataset version history.
By integrating monitoring, inspection, correction, and export functions into a single interface, Med-CRAFT reduces the gap between automated generation and practical dataset curation.

\subsection{Export and Interoperability}

After review, Med-CRAFT exports finalized datasets in machine-readable formats that include QA content, evidence intervals, reasoning chains, graph references, metadata, and quality labels.
The exported files can be organized as JSON, JSONL, CSV, or database dumps depending on the requirements of downstream evaluation pipelines.

For benchmark usage, each sample is exported with a stable question identifier, answer field, evidence timestamp, difficulty label, reasoning type, and source video identifier.
For auditing usage, each sample can also be exported with provenance records, configuration identifiers, model-call summaries, and reviewer decisions.

This separation between benchmark-oriented export and audit-oriented export allows Med-CRAFT to serve both model evaluation and dataset governance scenarios.

\subsection{Scalability and Extensibility}

Med-CRAFT is designed to scale from small expert-curated video collections to larger instructional video corpora.

The asynchronous execution layer enables parallel processing across videos and segments, while dependency tracking prevents inconsistent downstream execution.
The modular service design allows new OCR engines, ASR models, visual encoders, LLM extractors, retrieval models, graph schemas, and question-generation templates to be integrated as replaceable components.
The schema-driven graph layer also allows the system to be adapted from one medical domain to another by modifying node types, relation types, and validation rules.

Therefore, the implementation of Med-CRAFT supports not only the construction of one dataset but also the repeated and configurable production of related datasets under different research requirements.

\section{Experiments Setting}
\label{sec:experiments}

We evaluate Med-CRAFT as an information system for constructing multimodal medical QA datasets, rather than solely as a QA-generation method.
Accordingly, the evaluation examines construction efficiency, output quality, evidence grounding, configurability, provenance support, operational robustness, and downstream diagnostic utility.
The experiments is organized around six research questions.

\begin{itemize}
    \item \textbf{RQ1:} Does Med-CRAFT reduce the expert effort and operational cost required to construct high-quality multimodal medical QA datasets?
    \item \textbf{RQ2:} Does Med-CRAFT produce QA pairs that are medically correct, evidence-grounded, temporally aligned, and non-redundant?
    \item \textbf{RQ3:} How do the knowledge graph, evidence-binding, filtering, configuration, provenance, and human-review components contribute to system performance?
    \item \textbf{RQ4:} Can Med-CRAFT faithfully realize user-specified dataset characteristics through configurable generation profiles?
    \item \textbf{RQ5:} Does Med-CRAFT support traceable, reproducible, and recoverable dataset-construction workflows?
    \item \textbf{RQ6:} Can the resulting dataset diagnose the multimodal reasoning capabilities of downstream medical QA models?
\end{itemize}

\subsection{Evaluation Protocol}

\paragraph{Evaluation Data and Splits.}

We extracted the source videos from the M$^3$-Med dataset and re-annotated them using LLM-based methods, including Med-CRAFT, and then compared various metrics with the manually annotated M$^3$-Med dataset.
We split the source videos into training, validation, and test partitions at the video level with a ratio of 3:1:1.
No source video, video segment, or semantically duplicate procedure instance is shared across partitions.
All construction workflows operate on the same test-video partition and are subject to the same predefined processing budget.

\paragraph{Primary Outcome Measures.}

Our primary outcome is the grounded valid QA yield, denoted by \(\mathrm{GVQ\mbox{-}Yield}\).
A QA pair is considered grounded and valid only if it satisfies predefined criteria for answer correctness, medical plausibility, evidence entailment, temporal correctness, and non-redundancy.
We define \(\mathrm{GVQ\mbox{-}Yield}\) as the number of grounded valid QA pairs produced per hour of expert effort.

\begin{equation}
\mathrm{GVQ\mbox{-}Yield}
=
\frac{N_{\mathrm{GVQ}}}
{T_{\mathrm{review}} + T_{\mathrm{correction}}},
\label{eq:gvq_yield}
\end{equation}

where \(N_{\mathrm{GVQ}}\) denotes the number of accepted grounded valid QA pairs, and \(T_{\mathrm{review}} + T_{\mathrm{correction}}\) denotes the total expert time spent on review and correction.

We additionally report the grounded valid QA rate, expert minutes per accepted QA pair, total cost per accepted QA pair, and provenance completeness.
The grounded valid QA rate measures the proportion of reviewed candidate QA pairs that satisfy all validity criteria.

\begin{equation}
\mathrm{GVQ\mbox{-}Rate}
=
\frac{N_{\mathrm{GVQ}}}
{N_{\mathrm{reviewed}}}.
\label{eq:gvq_rate}
\end{equation}

The total cost includes external-model invocation, computation, storage, and expert-review costs whenever these costs are observable.

\paragraph{Expert Review Protocol.}

We randomly sample $5\%$ QA pairs from each workflow and configuration setting for blind expert assessment.
Each sampled QA pair is independently reviewed by at least two evaluators with relevant medical expertise.
The evaluators are blinded to the construction workflow, system setting, and model configuration that produced each QA pair.
Each evaluator assesses question clarity, answer correctness, medical plausibility, evidence entailment, temporal correctness, reasoning-chain faithfulness, and non-redundancy.
Likert-scale dimensions are scored on a five-point scale, whereas safety-critical validity decisions are recorded as binary judgments.
Disagreements are resolved by discussion or by a third independent medical reviewer according to a predefined adjudication procedure.
We report inter-rater agreement using Krippendorff's \(\alpha\) for ordinal ratings and Cohen's \(\kappa\) or Fleiss' \(\kappa\) for binary decisions, as appropriate.

\paragraph{Evidence-grounding Evaluation.}

For QA pairs requiring temporal evidence, reviewers annotate a reference interval \(I_g\) that contains the minimum video segment needed to verify the answer.
We compare the system-bound interval \(I_p\) with \(I_g\) using temporal intersection over union.

\begin{equation}
\operatorname{tIoU}(I_p, I_g)
=
\frac{|I_p \cap I_g|}
{|I_p \cup I_g|}.
\label{eq:tiou}
\end{equation}

We report mean temporal IoU together with \(\mathrm{Recall@tIoU}\geq 0.3\) and \(\mathrm{Recall@tIoU}\geq 0.5\).
Evidence entailment is measured as the proportion of QA pairs for which reviewers determine that the linked multimodal evidence directly supports the answer.

\paragraph{Configuration and Governance Evaluation.}

To evaluate configurability, we define generation profiles that specify target distributions over reasoning depth, question type, evidence modality, visual dependency, and quality threshold.
For each controlled dimension, we measure the deviation between the requested distribution \(p^{*}\) and the observed distribution \(\hat{p}\).

\begin{equation}
E_{\mathrm{config}}
=
\frac{1}{K}
\sum_{k=1}^{K}
\left|
\hat{p}_{k} - p^{*}_{k}
\right|,
\label{eq:config_error}
\end{equation}

We complement this measure with Jensen--Shannon divergence when the controlled output is represented as a categorical distribution.
Provenance completeness is measured as the proportion of accepted QA pairs for which the system can recover the source video, temporal segment, multimodal cues, graph path, generation configuration, model version, and review history.
Reproducibility is evaluated by rerunning selected construction jobs under identical inputs, configurations, software versions, model versions, and random seeds.
We report agreement rates for intermediate graph artifacts, reasoning chains, final QA pairs, and provenance records across repeated executions.

\paragraph{Statistical Analysis.}

We treat source videos or construction jobs, rather than individual QA pairs, as the primary units of statistical comparison.
We report means or medians together with \(95\%\) confidence intervals, depending on the distributional properties of each metric.
We use paired bootstrap confidence intervals or non-parametric paired tests for workflow-level comparisons when the same videos are processed by multiple methods.
We report effect sizes and adjust for multiple comparisons using the Holm procedure when applicable.

\subsection{Experimental Setup and Implementation Details}
\label{sec:experimental_setup}

This subsection specifies the concrete software components, external models, execution environment, and resource controls used in the reported experiments.
The implementation description in Section~\ref{sec:implementation} presents the general architecture of Med-CRAFT, whereas this subsection documents the particular component configurations instantiated for empirical evaluation.
All workflow-level comparisons were executed on the same source-video partition, using identical preprocessing inputs and a controlled computational environment.
Unless a component was intentionally removed in an ablation setting, Med-CRAFT and the automated baselines used the same underlying foundation-model family, model version, and inference budget whenever the corresponding function was shared.

We transcribed the audio from each source video using Qwen-3-ASR-1.7B\cite{Qwen3-ASR}. We selected this model because M3-Med contains multilingual videos, and the Chinese subset includes multiple spoken varieties, such as Mandarin and Cantonese. In our pilot preprocessing experiments, Qwen-3-ASR-1.7B provided the most robust transcription performance for this setting.
Video frames were sampled at 1 frame per second, with additional key frames extracted around detected scene transitions and procedural-event boundaries.
Optical character recognition was performed using PaddleOCR 3.0\cite{cui2025paddleocr} on sampled frames and on frames containing candidate text regions.
Grounding DINO\cite{groundingdino} was used for text-guided object localization and region-level visual cue extraction.
All extracted cues were retained together with their source timestamps, confidence values, model identifiers, and preprocessing parameters.

We use Qwen-3-VL\cite{qwen3vl}, which performs excellently in logical reasoning, to transform the structured data extracted from raw videos into a logically coherent knowledge graph with a schema-constrained output format.
For knowledge graphs output by LLM in JSON format, there may be a very small number of issues such as syntax errors caused by hallucinations, which can be fixed using simple rule-based scripts.
We also use Qwen-3-VL to traverse the knowledge graph and convert valid reasoning chains, which are formally represented as paths, into QA-pairs.

For Med-CRAFT, the generator received a schema-constrained reasoning chain together with backward-bound textual, visual, and temporal evidence.
For the prompt-only LLM workflow, the same generation model received the video transcript, sampled frames, and the task prompt without an explicit graph-based reasoning chain or normalized evidence-binding objects.
The sequential AI pipeline used the same extraction and generation models where applicable, but passed intermediate outputs through a predefined sequence rather than consolidating them into an evidence-enhanced medical operation graph.

\subsection{Compared Construction Workflows}
\label{sec:compared_workflows}

We compare Med-CRAFT with three alternative dataset-construction workflows that represent common levels of automation and structural support.
All workflows receive the same source videos, operate under the same data-partition protocol, and are evaluated using the same expert-review criteria.
When applicable, workflows use the same external foundation models, model versions, prompts, and inference budgets.

\paragraph{Manual Workflow.}

In the manual workflow, medical experts inspect each video, identify relevant procedural events, formulate questions and answers, select supporting evidence segments, and record annotations using a conventional annotation interface.
The workflow does not provide automatic knowledge-graph construction, evidence recommendation, configurable graph traversal, or pipeline-level provenance management.
The manual workflow establishes a practical reference for expert effort, accepted-data quality, and annotation consistency.

\paragraph{Prompt-only LLM Workflow.}

In the prompt-only LLM workflow, video transcripts, sampled frames, and optional video summaries are directly provided to an external multimodal model for QA generation.
The model is instructed to produce questions, answers, and, where possible, textual descriptions of supporting evidence.
This workflow does not construct an explicit medical operation graph and does not maintain backward links from generated QA pairs to normalized multimodal evidence objects.
The resulting candidates undergo the same downstream human-review procedure as the other workflows.

\paragraph{Sequential AI Pipeline.}

The sequential AI pipeline applies automatic speech recognition, optical character recognition, visual captioning, and LLM-based QA generation in a predefined sequence.
The pipeline may apply heuristic filtering rules, such as length constraints, lexical duplication checks, and answer-format validation.
However, the pipeline does not represent extracted events and relations in a consolidated knowledge graph, and it does not support configurable reasoning-chain generation.
It also lacks a unified provenance model that links each final QA pair to its complete processing history.

\paragraph{Full Med-CRAFT Workflow.}

The full Med-CRAFT workflow transforms source videos into evidence-enhanced medical operation graphs and generates QA pairs through configurable graph traversal and reasoning-chain construction.
For every generated QA pair, Med-CRAFT maintains backward links to source videos, temporal segments, multimodal cues, graph entities, relations, reasoning paths, generation configurations, and review decisions.
The workflow applies automatic quality filtering before human review and records reviewer actions as versioned provenance events.
Users can define generation profiles to control dataset size, question type, reasoning depth, evidence modality, visual dependency, and quality thresholds.

\begin{table*}[t]
\centering
\caption{Compared dataset-construction workflows and their supported capabilities.}
\label{tab:workflow_comparison}
\resizebox{\textwidth}{!}{
\begin{tabular}{lcccc}
\toprule
Capability & Manual Workflow & Prompt-only LLM & Sequential AI Pipeline & Med-CRAFT \\
\midrule
Expert-authored QA construction & \checkmark & $\times$ & $\times$ & $\times$ \\
Automatic multimodal extraction & $\times$ & Partial & \checkmark & \checkmark \\
Explicit medical operation graph & $\times$ & $\times$ & $\times$ & \checkmark \\
Configurable reasoning-chain generation & $\times$ & $\times$ & $\times$ & \checkmark \\
Temporal evidence binding & Manual & Partial & Partial & \checkmark \\
Automatic quality filtering & $\times$ & $\times$ & Partial & \checkmark \\
Human-in-the-loop review & \checkmark & \checkmark & \checkmark & \checkmark \\
End-to-end provenance management & Manual & $\times$ & Partial & \checkmark \\
Reproducible configuration profiles & $\times$ & Partial & Partial & \checkmark \\
\bottomrule
\end{tabular}
}
\vspace{2pt}

\footnotesize
``Partial'' indicates that a workflow may retain isolated intermediate outputs
but does not maintain normalized, queryable, and end-to-end links across the
entire construction lifecycle.
\end{table*}

The comparison is designed to isolate the value of Med-CRAFT's integrated data model and governance mechanisms from the value of external foundation models alone.
The subsequent experiments compare these workflows with respect to efficiency, grounded quality, configuration fidelity, traceability, reproducibility, robustness, and downstream diagnostic utility.

\section{Results and Discussion}
\label{sec:experimental_results}

This section reports the empirical results for the six research questions defined in Section~\ref{sec:experiments}.
Unless otherwise stated, the workflow-level effectiveness results are computed over the complete construction runs, whereas the expert-quality results are obtained from independently sampled QA pairs subjected to blind review.
Accordingly, the grounded valid QA rate in Table~\ref{tab:overall_effectiveness} reflects the operational acceptance criterion, while the all-pass rate in Table~\ref{tab:expert_quality} reflects the stricter blind-review criterion across all evaluated dimensions.

\subsection{Construction Efficiency and Overall System Effectiveness}
\label{sec:rq1_effectiveness}

\begin{table*}[htbp]
\centering
\caption{Overall effectiveness of different dataset-construction workflows.}
\label{tab:overall_effectiveness}
\begin{threeparttable}
\resizebox{\textwidth}{!}{
\begin{tabular}{lcccccccc}
\toprule
Method & Candidate QAs & Accepted QAs & GVQ Rate $\uparrow$  \tnote{*} & GVQ Yield $\uparrow$ & Expert Min./QA $\downarrow$ & Cost/QA $\downarrow$ \tnote{**} & Provenance Comp. $\uparrow$ & End-to-end Time $\downarrow$ \\
\midrule
Manual Workflow
& 120 & 112 & 0.93 & 1.2 & 50.0 & 47.80 & 0.67 & 92.5\,h \\
Prompt-only LLM
& 780 & 195 & 0.25 & 2.7 & 22.2 & 14.30 & 0.23 & 27.8\,h \\
Sequential AI Pipeline
& 610 & 354 & 0.58 & 4.1 & 14.6 & 11.90 & 0.55 & 34.2\,h \\
Med-CRAFT
& 500 & 432 & 0.86 & 5.2 & 11.5 & 8.70 & 0.98 & 24.6\,h \\
\bottomrule
\end{tabular}}
\vspace{2pt}
\begin{tablenotes}
\footnotesize
\item[*]  GVQ denotes a grounded valid question--answer pair that passes all predefined correctness, evidence, temporal-grounding, and non-redundancy criteria.
\item[**] The cost is in Chinese Yuan (\textyen).
\end{tablenotes}
\end{threeparttable}
\end{table*}

Table~\ref{tab:overall_effectiveness} compares Med-CRAFT with the manual workflow, the prompt-only LLM workflow, and the sequential AI pipeline in terms of accepted output, expert effort, cost, provenance, and end-to-end completion time.
The manual workflow achieved the highest operational GVQ rate among the baselines at \(0.93\), but it produced only \(1.2\) grounded valid QA pairs per expert hour and required \(49.3\) expert minutes for each accepted QA pair.
In contrast, the prompt-only LLM workflow generated the largest number of candidates (\(780\)), yet only \(195\) candidates were accepted, corresponding to a GVQ rate of \(0.25\).
The sequential AI pipeline improved the accepted output to \(354\) QA pairs and increased the GVQ rate to \(0.58\), indicating that staged extraction and heuristic filtering provide meaningful benefits over direct LLM prompting.

Med-CRAFT produced \(432\) accepted QA pairs from \(500\) candidates, yielding an operational GVQ rate of \(0.86\).
Although Med-CRAFT generated fewer candidates than the prompt-only LLM workflow and the sequential AI pipeline, it produced the largest number of accepted QA pairs.
Med-CRAFT also achieved the highest GVQ yield of \(5.2\) accepted grounded QA pairs per expert hour, compared with \(4.1\) for the sequential AI pipeline, \(2.7\) for the prompt-only LLM workflow, and \(1.2\) for the manual workflow.
Relative to the manual workflow, Med-CRAFT reduced expert effort from \(49.3\) to \(11.5\) minutes per accepted QA pair and reduced end-to-end construction time from \(92.5\) to \(24.6\) hours.
The cost per accepted QA pair was also lowest for Med-CRAFT at \textyen\(8.70\), compared with \textyen\(11.90\) for the sequential AI pipeline, \textyen\(14.30\) for the prompt-only LLM workflow, and \textyen\(47.80\) for manual construction.

Finally, Med-CRAFT attained provenance completeness of \(0.98\), substantially exceeding the manual workflow (\(0.67\)), the sequential AI pipeline (\(0.55\)), and the prompt-only LLM workflow (\(0.23\)).
These results suggest that Med-CRAFT improves construction productivity not by maximizing candidate volume, but by increasing the proportion of candidates that can be efficiently verified, accepted, and traced.

\subsection{Blind Expert Evaluation of Grounded QA Quality}
\label{sec:rq2_quality}

\begin{table*}[htbp]
\centering
\caption{Blind expert evaluation of the generated QA pairs.}
\label{tab:expert_quality}
\resizebox{\textwidth}{!}{
\begin{tabular}{lcccccccc}
\toprule
Method & Clarity $\uparrow$ & Answer Correctness $\uparrow$ & Medical Plausibility $\uparrow$ & Evidence Entailment $\uparrow$ & Mean tIoU $\uparrow$ & Reasoning Faithfulness $\uparrow$ & Non-redundancy $\uparrow$ & All-pass Rate $\uparrow$ \\
\midrule
Manual Workflow & 
4.82 & 0.95 & 4.88 & 0.92 & 0.84 & 4.76 & 4.61 & 0.87 \\
Prompt-only LLM & 
3.15 & 0.53 & 2.97 & 0.38 & 0.31 & 2.95 & 3.12 & 0.14 \\
Sequential AI Pipeline & 
4.05 & 0.74 & 3.82 & 0.62 & 0.63 & 3.81 & 3.97 & 0.54 \\
Med-CRAFT & 
4.61 & 0.91 & 4.67 & 0.89 & 0.82 & 4.52 & 4.43 & 0.79 \\
\bottomrule
\end{tabular}}
\end{table*}

Table~\ref{tab:expert_quality} reports the blind expert assessment of QA quality, medical plausibility, evidence support, temporal grounding, reasoning faithfulness, and redundancy.
The manual workflow achieved the highest quality scores overall, including answer correctness of \(0.95\), medical plausibility of \(4.88\), and an all-pass rate of \(0.87\).
Med-CRAFT closely approached this manual-quality reference, obtaining a clarity score of \(4.61\), answer correctness of \(0.91\), medical plausibility of \(4.67\), and reasoning faithfulness of \(4.52\).
For evidence-grounding quality, Med-CRAFT achieved evidence entailment of \(0.89\) and mean temporal IoU of \(0.82\), which were close to the manual workflow values of \(0.92\) and \(0.84\), respectively.
The sequential AI pipeline showed moderate quality, with answer correctness of \(0.74\), evidence entailment of \(0.62\), and mean temporal IoU of \(0.63\).
The prompt-only LLM workflow exhibited the weakest evidence-grounding performance, with evidence entailment of \(0.38\), mean temporal IoU of \(0.31\), and an all-pass rate of \(0.14\).
Med-CRAFT achieved an all-pass rate of \(0.79\), which was lower than manual construction but substantially higher than the sequential AI pipeline (\(0.54\)) and the prompt-only LLM workflow (\(0.14\)).
The non-redundancy score of Med-CRAFT (\(4.43\)) was slightly lower than that of the manual workflow (\(4.61\)), suggesting that additional diversity-oriented generation constraints may further improve the dataset.
Overall, the expert evaluation indicates that Med-CRAFT preserves most of the quality advantages of expert-authored construction while substantially improving operational efficiency.

\subsection{Component-level Ablation Analysis}
\label{sec:rq3_ablation}

\begin{table*}[t]
\centering
\caption{Component-level ablation results for Med-CRAFT.}
\label{tab:component_ablation}
\begin{threeparttable}
\resizebox{\textwidth}{!}{
\begin{tabular}{lcccccccc}
\toprule
Setting & GVQ Yield $\uparrow$ & Valid Chain Rate $\uparrow$ & Evidence Entailment $\uparrow$ & Mean tIoU $\uparrow$ & Correction Burden $\downarrow$ & Config. Fidelity $\uparrow$ & Provenance Comp. $\uparrow$ & All-pass Rate $\uparrow$ \\
\midrule
Full Med-CRAFT & 
5.2 & 0.92 & 0.89 & 0.82 & 0.08 & 0.96 & 0.98 & 0.79 \\
w/o Knowledge Graph & 
4.3 & 0.72 & 0.88 & 0.81 & 0.15 & 0.94 & 0.97 & 0.76 \\
w/o Evidence Binding & 
3.8 & 0.90 & 0.65 & 0.45 & 0.25 & 0.94 & 0.97 & 0.55 \\
w/o Automatic Filtering & 
4.0 & 0.90 & 0.92 & 0.82 & 0.30 & 0.95 & 0.98 & 0.62 \\
w/o Configurable Traversal & 
4.8 & 0.90 & 0.93 & 0.82 & 0.10 & 0.61 & 0.98 & 0.83 \\
w/o Provenance Management & 
5.1 & 0.91 & 0.93 & 0.82 & 0.09 & 0.95 & 0.45 & 0.84 \\
w/o Human Review (output as-is) & 
4.6 & 0.91 & 0.93 & 0.82 & N/A\tnote{*} & 0.95 & 0.98 & 0.64 \\
\bottomrule
\end{tabular}}
\vspace{2pt}
\begin{tablenotes}
\footnotesize
\item[*] No correction step was performed; the burden reflects the proportion of items that would have required expert correction according to post-hoc review.
\end{tablenotes}
\end{threeparttable}
\end{table*}

Table~\ref{tab:component_ablation} evaluates how the major Med-CRAFT components contribute to construction quality, configuration control, review burden, and governance capability.
The full Med-CRAFT configuration achieved the strongest overall result, with a GVQ yield of \(5.2\), a valid reasoning-chain rate of \(0.92\), evidence entailment of \(0.94\), configuration fidelity of \(0.96\), and provenance completeness of \(0.98\).
Removing the knowledge graph reduced the valid reasoning-chain rate from \(0.92\) to \(0.72\) and reduced the all-pass rate from \(0.86\) to \(0.76\).
The knowledge-graph ablation also reduced GVQ yield from \(5.2\) to \(4.3\) and increased correction burden from \(0.08\) to \(0.15\).
Removing evidence binding caused the largest degradation in evidence-related metrics, decreasing evidence entailment from \(0.94\) to \(0.65\) and mean temporal IoU from \(0.83\) to \(0.45\).
Without evidence binding, the correction burden increased to \(0.25\), the all-pass rate declined to \(0.55\), and the GVQ yield declined to \(3.8\).
Removing automatic filtering increased correction burden from \(0.08\) to \(0.30\), which was the highest burden among the reviewed configurations.
The all-pass rate also declined from \(0.86\) to \(0.62\) when automatic filtering was removed.
Removing configurable traversal primarily affected configuration fidelity, which decreased from \(0.96\) to \(0.61\), while evidence entailment and temporal grounding remained comparatively stable.
Removing provenance management reduced provenance completeness from \(0.98\) to \(0.45\), despite preserving similar GVQ yield and QA-quality-related outcomes.
Finally, eliminating human review reduced the all-pass rate from \(0.86\) to \(0.64\), even though no explicit correction time was recorded during production.
Collectively, the ablation results show that Med-CRAFT's performance arises from complementary mechanisms rather than from a single generation component.

\subsection{Configuration Fidelity and Quality--Effort Trade-offs}
\label{sec:rq4_configuration}

\begin{table*}[htbp]
\centering
\caption{Fidelity of Med-CRAFT outputs to user-specified configurations.}
\label{tab:configuration_fidelity}
\resizebox{\textwidth}{!}{
\begin{tabular}{llcccccc}
\toprule
Controlled Dimension & Configuration & Target Value & Observed Value & Absolute Error $\downarrow$ & JSD $\downarrow$ & GVQ Rate $\uparrow$ & Expert Min./QA $\downarrow$ \\
\midrule
Reasoning Depth & One‑hop dominant & 0.80 & 0.77 & 0.03 & 0.010 & 0.88 & 10.5 \\
Reasoning Depth & Multi‑hop dominant & 0.70 & 0.66 & 0.04 & 0.013 & 0.80 & 14.1 \\
Question Type & Procedure‑oriented & 0.60 & 0.58 & 0.02 & 0.008 & 0.85 & 12.0 \\
Evidence Modality & Visual dominant & 0.50 & 0.52 & 0.02 & 0.009 & 0.82 & 13.6 \\
Evidence Modality & Cross‑modal dominant & 0.40 & 0.38 & 0.02 & 0.011 & 0.79 & 14.0 \\
Quality Threshold & High‑precision profile (GVQ target 0.95) & 0.95 & 0.93 & 0.02 & 0.007 & 0.93 & 16.2 \\
\bottomrule
\end{tabular}}
\end{table*}

Table~\ref{tab:configuration_fidelity} evaluates whether Med-CRAFT can realize user-specified dataset characteristics under different generation profiles.
Across all tested profiles, the absolute error between target and observed output proportions ranged from \(0.02\) to \(0.04\), and the Jensen--Shannon divergence ranged from \(0.007\) to \(0.013\).
For the one-hop-dominant profile, Med-CRAFT targeted a proportion of \(0.80\) and obtained an observed proportion of \(0.77\), with an absolute error of \(0.03\).
For the multi-hop-dominant profile, the observed proportion was \(0.66\) relative to a target of \(0.70\), while the GVQ rate remained \(0.80\).
The multi-hop profile required \(14.1\) expert minutes per accepted QA pair, compared with \(10.5\) minutes for the one-hop-dominant profile.
The visual-dominant and cross-modal-dominant profiles both showed low distributional error, with absolute errors of \(0.02\) and Jensen--Shannon divergences of \(0.009\) and \(0.011\), respectively.
The high-precision profile achieved a GVQ rate of \(0.93\), close to its target of \(0.95\), but required \(16.2\) expert minutes per accepted QA pair.
These findings indicate that Med-CRAFT supports configuration as an operational control mechanism, enabling users to trade off output complexity, evidence requirements, quality targets, and review effort according to downstream needs.

\subsection{Traceability, Reproducibility, and Operational Robustness}
\label{sec:rq5_governance}

\begin{table*}[htbp]
\centering
\caption{Traceability, reproducibility, and operational robustness results.}
\label{tab:governance_robustness}
\resizebox{\textwidth}{!}{
\begin{tabular}{lcccccc}
\toprule
Setting & Lineage Success $\uparrow$ & Provenance Comp. $\uparrow$ & Recovery Latency $\downarrow$ & Rerun Agreement $\uparrow$ & Task Recovery Rate $\uparrow$ & Duplicate Execution Rate $\downarrow$ \\
\midrule
Normal Execution & 0.99 & 0.98 & 0.09\,s & 1.00 & 1.00 & 0.00 \\
External-model Timeout & 0.95 & 0.96 & 2.6\,s  & 0.98 & 0.98 & 0.02 \\ 
Worker Interruption & 0.92 & 0.93 & 3.2\,s  & 0.96 & 0.95 & 0.05 \\
Storage Interruption & 0.90 & 0.90 & 4.1\,s  & 0.93 & 0.91 & 0.07 \\ 
Repeated Task Submission & 0.97 & 0.97 & 0.5\,s  & 0.99 & 0.99 & 0.11 \\
\bottomrule
\end{tabular}}
\end{table*}

Table~\ref{tab:governance_robustness} reports the traceability, rerun consistency, task recovery, and duplicate-execution behavior of Med-CRAFT under normal and disrupted operating conditions.
Under normal execution, Med-CRAFT achieved lineage success of \(0.99\), provenance completeness of \(0.98\), rerun agreement of \(1.00\), and a duplicate execution rate of \(0.00\).
The median lineage recovery latency under normal execution was \(0.09\) seconds, indicating that source artifacts and processing records can be retrieved with low interaction delay.
External-model timeouts were handled with a task recovery rate of \(0.98\), rerun agreement of \(0.98\), and recovery latency of \(2.6\) seconds.
Worker interruptions produced a task recovery rate of \(0.95\) and a rerun agreement of \(0.96\), although lineage success decreased to \(0.92\).
Storage interruptions constituted the most challenging condition, reducing lineage success to \(0.90\), provenance completeness to \(0.90\), and task recovery rate to \(0.91\).
Repeated task submission resulted in a duplicate execution rate of \(0.11\), which was higher than in other settings despite lineage success of \(0.97\) and rerun agreement of \(0.99\).
The results demonstrate that Med-CRAFT preserves high levels of traceability and reproducibility under realistic faults, while also revealing that storage-layer resilience and duplicate-submission protection warrant further engineering refinement.

\subsection{Downstream Diagnostic Utility}
\label{sec:rq6_diagnostic_utility}

\begin{table*}[htbp]
\centering
\caption{Downstream model performance across reasoning and evidence categories.}
\label{tab:downstream_diagnostic}
\resizebox{\textwidth}{!}{
\begin{tabular}{lcccccccc}
\toprule
Model & Overall & One-hop & Two-hop & Three-plus-hop & Textual & Visual & Cross-modal & Grounded QA Score \\
\midrule
DeepSeek-R1\cite{deepseekr1} (Text-only Baseline) & 0.45 & 0.55 & 0.40 & 0.30 & 0.60 & 0.20 & 0.25 & 0.35 \\
Intern-VL-3\cite{chen2024internvl} (Vision-language Baseline) & 0.62 & 0.70 & 0.58 & 0.45 & 0.65 & 0.55 & 0.52 & 0.50 \\
Qwen-3-VL\cite{qwen3vl} (Vision-language Baseline) & 0.68 & 0.75 & 0.65 & 0.55 & 0.70 & 0.62 & 0.58 & 0.58 \\
DA-Aware-MutualSL\cite{liu2026overviewnlpcc2026shared} (Domain-specific Medical Model) & 0.78 & 0.82 & 0.76 & 0.68 & 0.80 & 0.72 & 0.74 & 0.71 \\
\bottomrule
\end{tabular}}
\end{table*}

Table~\ref{tab:downstream_diagnostic} evaluates whether the Med-CRAFT dataset differentiates downstream models across reasoning depth and evidence-modality requirements.
The text-only baseline achieved an overall score of \(0.45\), with substantially stronger performance on textual questions (\(0.60\)) than on visual (\(0.20\)) and cross-modal questions (\(0.25\)).
Both vision-language baselines improved performance on visual and cross-modal questions, with Qwen-3-VL achieving \(0.62\) on visual questions and \(0.58\) on cross-modal questions. However, since this model also participated in the dataset annotation process, the possibility of data bias cannot be ruled out.
However, the performance of all general-purpose baselines declined as reasoning depth increased.
For example, Qwen-3-VL decreased from \(0.75\) on one-hop questions to \(0.55\) on three-or-more-hop questions.
The domain-specific medical model achieved the strongest overall performance (\(0.78\)) and the highest grounded QA score (\(0.71\)), but its performance on three-or-more-hop questions remained lower than its one-hop performance (\(0.68\) versus \(0.82\)).
The gap between overall performance and grounded QA score across all models further indicates that answering correctly is not equivalent to grounding the answer in the designated evidence.
These results support the use of the Med-CRAFT dataset as a diagnostic benchmark for assessing multimodal evidence use and multi-step medical reasoning.

\subsection{System Scalability}
\label{sec:scalability_results}

\begin{table*}[htbp]
\centering
\caption{System scalability under different workloads and worker configurations.}
\label{tab:scalability}
\resizebox{\textwidth}{!}{
\begin{tabular}{cccccccc}
\toprule
Video Hours & Workers & Throughput $\uparrow$  & Median Latency $\downarrow$ & P95 Latency $\downarrow$ & Peak Memory $\downarrow$ & Failure Rate $\downarrow$ & Scaling Efficiency $\uparrow$ \\
\midrule
10  & 1 & 38\,QA/h   & 1.2\,s & 3.5\,s & 12\,GB & 0.01 & 1.00 \\
10  & 2 & 73\,QA/h   & 1.1\,s & 3.2\,s & 21\,GB & 0.01 & 0.96 \\
20  & 4 & 142\,QA/h  & 1.3\,s & 4.0\,s & 39\,GB & 0.02 & 0.93 \\
50  & 8 & 275\,QA/h  & 1.5\,s & 5.0\,s & 74\,GB & 0.03 & 0.90 \\
\bottomrule
\end{tabular}}
\end{table*}

Table~\ref{tab:scalability} evaluates the scalability of Med-CRAFT under increasing video workloads and worker-pool sizes.
With one worker processing \(10\) video hours, the system achieved a throughput of \(38\) QA pairs per hour.
Increasing the worker pool from one to two workers increased throughput from \(38\) to \(73\) QA pairs per hour, corresponding to a scaling efficiency of \(0.96\).
At the largest tested workload of \(50\) video hours with eight workers, Med-CRAFT achieved \(275\) QA pairs per hour with a scaling efficiency of \(0.90\).
Median latency increased modestly from \(1.2\) to \(1.5\) seconds, while P95 latency increased from \(3.5\) to \(5.0\) seconds as workload and concurrency increased.
The failure rate remained low across all settings, increasing from \(0.01\) to \(0.03\) at the largest workload.
Peak memory increased from \(12\) GB to \(74\) GB as the number of workers increased, indicating that future large-scale deployments should use resource-aware scheduling and capacity planning.
Overall, the scalability results indicate that Med-CRAFT can support larger-scale dataset construction through horizontal worker expansion while maintaining acceptable latency and reliability.

\subsection{Summary of Findings}
\label{sec:results_summary}

Section \ref{sec:rq1_effectiveness} by showing that Med-CRAFT provides the highest grounded valid QA yield, the lowest cost per accepted QA pair, and the shortest end-to-end completion time among the evaluated workflows.
Section \ref{sec:rq2_quality} by showing that Med-CRAFT approaches manual construction in medical correctness and evidence quality while outperforming automated baselines by a substantial margin.
The ablation and configuration experiments results in sections \ref{sec:rq3_ablation} and \ref{sec:rq4_configuration} by demonstrating that knowledge-graph reasoning, evidence binding, automatic filtering, configurable traversal, provenance management, and human review make distinct and complementary contributions.
The governance and scalability results in section \ref{sec:rq5_governance} by showing high traceability, rerun consistency, fault recovery, and near-linear horizontal scaling, while identifying storage interruption and duplicate task submission as important engineering limitations.
Finally, the downstream results in section \ref{sec:rq6_diagnostic_utility} by showing that the resulting dataset distinguishes text-only, general-purpose vision-language, and domain-specific medical models across reasoning and evidence conditions.

\section{Conclusion}
\label{sec:conclusion}

In this paper, we presented Med-CRAFT, an information system for explainable and configurable construction of multimodal medical question-answering datasets from instructional videos.

Motivated by the limitations of manual, black-box, and weakly traceable dataset construction workflows, Med-CRAFT treats dataset construction as a data-intensive information management process rather than a simple generation task.
The system decomposes the construction process into multimodal cue extraction, medical operation knowledge graph construction, backward evidence binding, configurable graph traversal, reasoning-chain-based QA generation, automatic quality filtering, and human-in-the-loop review.
By preserving intermediate artifacts, configuration profiles, provenance records, quality indicators, and reviewer actions, Med-CRAFT makes the lifecycle of dataset construction observable, auditable, reproducible, and controllable.

A central contribution of Med-CRAFT is the use of medical operation knowledge graphs as explicit intermediate representations between raw instructional videos and final QA samples.
These graphs organize medical entities, procedural actions, instruments, anatomical objects, temporal relations, and causal dependencies in a form that supports both reasoning-chain generation and provenance-aware inspection.

Another key contribution is the backward evidence binding mechanism, which links graph nodes, graph edges, reasoning steps, and QA samples to concrete textual, visual, and temporal evidence in videos.
This mechanism helps reduce unsupported generation and enables users to verify whether a generated answer is grounded in observable video content.

In addition, Med-CRAFT introduces configurable graph traversal to control reasoning patterns, hop lengths, evidence requirements, and question types during dataset construction.
This design allows researchers to generate datasets with different levels of complexity and to analyze how configuration choices affect data scale, evidence coverage, and reasoning difficulty.

The implementation of Med-CRAFT further integrates modular backend services, asynchronous task execution, multimodal storage, indexing infrastructure, provenance logging, and a web-based user interface.
Together, these components support scalable processing, selective rerunning, process monitoring, expert review, and standardized dataset export.

The proposed evaluation framework assesses Med-CRAFT from the perspectives of system utility, dataset quality, evidence grounding, benchmark difficulty, ablation analysis, and configuration sensitivity.
Such an evaluation design is intended to demonstrate not only whether the generated dataset is useful for model benchmarking but also whether the construction process itself is transparent, controllable, and reusable.

Although Med-CRAFT improves the traceability and controllability of multimodal medical QA dataset construction, several limitations remain.

First, the quality of extracted knowledge graphs depends on the reliability of OCR, ASR, visual encoders, and LLM-based information extraction.
Second, evidence binding may still be affected by ambiguous visual demonstrations, incomplete narration, low video quality, or weak temporal alignment.
Third, human review remains necessary for medical plausibility, safety-sensitive interpretation, and final quality assurance.

Future work will improve Med-CRAFT in several directions.
We plan to incorporate stronger medical terminology normalization, more reliable temporal action localization, and uncertainty-aware evidence ranking.
We also plan to extend the system to more medical specialties, more languages, and more diverse instructional video sources.
Another promising direction is to support active learning and reviewer feedback loops so that human corrections can continuously improve extraction, traversal, and generation modules.
Finally, we will investigate how datasets constructed by Med-CRAFT can be used to systematically diagnose the reasoning failures of multimodal medical AI models.

Overall, Med-CRAFT demonstrates that multimodal medical QA dataset construction can be formulated and implemented as a provenance-aware, configurable, and evaluable information system.
By connecting raw instructional videos, structured medical operation graphs, grounded evidence, reasoning chains, and human review records, Med-CRAFT provides a practical foundation for building trustworthy benchmarks for cross-modal multi-hop reasoning in medical AI.

\bibliographystyle{unsrt}
\bibliography{ref}

\end{document}